\documentclass{article}

\PassOptionsToPackage{numbers, compress}{natbib}
\usepackage{natbib}

\usepackage[preprint]{neurips_2026}

\makeatletter
\renewcommand{\@noticestring}{}
\makeatother

\usepackage[utf8]{inputenc} 
\usepackage[T1]{fontenc}    
\usepackage{hyperref}       
\usepackage{url}            
\usepackage{booktabs}       
\usepackage{amsfonts}       
\usepackage{nicefrac}       
\usepackage{microtype}      
\usepackage{xcolor}         

\usepackage{tocloft}
\usepackage{algorithm}
\usepackage{algorithmic}
\usepackage{graphicx}
\usepackage{hyperref}
\usepackage{amsmath}
\usepackage{amssymb}
\usepackage{mathtools}
\usepackage{amsthm}
\usepackage[capitalize,noabbrev]{cleveref}
\theoremstyle{plain}
\newtheorem{theorem}{Theorem}[section]
\newtheorem{proposition}[theorem]{Proposition}
\newtheorem{lemma}[theorem]{Lemma}
\newtheorem{corollary}[theorem]{Corollary}
\theoremstyle{definition}
\newtheorem{definition}[theorem]{Definition}

\theoremstyle{remark}
\newtheorem{remark}[theorem]{Remark}
\usepackage[textsize=tiny]{todonotes}
\usepackage{subcaption}

\usepackage{array}

\title{Performance–Efficiency Tradeoffs in Transformers: \\An Approximation Theory Perspective}

\author{
\begin{tabular}{c@{\hspace{1.6em}}c@{\hspace{1.6em}}c}
\textbf{Ruoxi Yu\textsuperscript{1,*}}
&
\textbf{Haotian Jiang\textsuperscript{2,*}}
&
\textbf{Jingpu Cheng\textsuperscript{3}}
\\[-0.15em]
{\scriptsize\texttt{yuruoxi@stu.pku.edu.cn}}
&
{\scriptsize\texttt{haotian@nus.edu.sg}}
&
{\scriptsize\texttt{chengjingpu@u.nus.edu}}
\\[0.55em]
\textbf{Penghao Yu\textsuperscript{3}}
&
\textbf{Qianxiao Li\textsuperscript{2,3}}
&
\textbf{Zhong Li\textsuperscript{4}}
\\[-0.15em]
{\scriptsize\texttt{e1353366@u.nus.edu}}
&
{\scriptsize\texttt{qianxiao@nus.edu.sg}}
&
{\scriptsize\texttt{lzhong@microsoft.com}}
\\[0.65em]
\multicolumn{3}{c}{
{\small \textsuperscript{1}Center for Data Science, Peking University}
}
\\[-0.05em]
\multicolumn{3}{c}{
{\small \textsuperscript{2}Institute for Functional Intelligent Materials,
National University of Singapore}
}
\\[-0.05em]
\multicolumn{3}{c}{
{\small \textsuperscript{3}Department of Mathematics,
National University of Singapore}
}
\\[-0.05em]
\multicolumn{3}{c}{
{\small \textsuperscript{4}Microsoft Research Asia}
}
\\[0.25em]
\multicolumn{3}{c}{
{\footnotesize \textsuperscript{*}Equal contribution.}
}
\end{tabular}
}

\begin{document}

\maketitle

\begin{abstract}
  Transformers have achieved remarkable empirical successes across a wide range of applications, yet how to balance model performance and computation efficiency in a principled manner remains underexplored. In this work, we theoretically study the parameter efficiency in multi-head attention Transformers by quantitatively analyzing the relationship between Transformers' hyper-parameters (i.e. the number of attention heads and head dimensions) and performance. (i) With a fixed total parameters budget, we establish an approximation error estimate for information extraction tasks, revealing an intrinsic trade-off between the number of heads and head dimensions to boost model performance. (ii) With maintained performance, we further mathematically prove a saturation pattern in softmax-based attention, showing that increasing head dimensions leads to diminishing returns in performance (especially for long sequences), implying potential parameter reduction. 
  All these theoretical results are consistently verified through extensive experiments from synthetic to real-world simulations, providing principled guidance for trade-offs between performance and parameter efficiency of Transformers. 
\end{abstract}

\section{Introduction}
\label{introduction}

Transformers have achieved remarkable successes across numerous applications such as natural language processing (\cite{devlin2019bert,brown2020language,grattafiori2024llama}), computer vision (\cite{dosovitskiy2021image,liu2021swin,wu2021cvt}), audio/speech (\cite{gong2022ssast, zhang2021transmask, lu2021spectnt, latif2023transformers}) and beyond, serving as the backbone of modern large language models (LLMs) and multi-modal systems due to their ability to capture long-range dependencies and scale efficiently (\cite{vaswani2017attention, dao2022flashattention, choromanski2020rethinking, wang2020linformer, beltagy2020longformer}). Their impact spans healthcare (\cite{Meng2024, Busch2025}), robotics (\cite{Zeng2023, Liu2025}), agent (\cite{shinn2023reflexionlanguageagentsverbal,yao2023reactsynergizingreasoningacting}), chemistry (\cite{Bran2024, Jablonka2024}), finance (\cite{zhou2021informer, wu2021autoformer}), and so on.

Despite their success, large language models face a fundamental efficiency challenge, as naive scaling demands rapidly growing data and computation (\cite{sorscher2022beyond,bartoldson2023compute}). A growing body of work has investigated improving Transformer efficiency through parameter allocation and architectural design.
In particular, multi-head attention (MHA) has been shown to exhibit substantial redundancy: Many attention heads can be pruned without significant performance drops, and only a small subset consistently captures meaningful patterns across tasks and languages (\cite{michel2019sixteen,voita2019analyzing,ma2021contributions}).
Building on these observations, subsequent studies propose automated and stable pruning criteria for model compression while preserving model performance (\cite{choi2025entropy,lee2024automatic}).
Meanwhile, computation efficiency has been further improved through architecture and system-level modifications, including sparse or hierarchical attention mechanisms (\cite{liu2021swin,chu2021twins,hassani2023neighborhood,beltagy2020longformer,zaheer2020bigbird}), shared key–value projections (\cite{ainslie2023gqa}), and optimized attention kernels (\cite{dao2022flashattention,hatamizadeh2023fastervit}).
Despite that all these prior advances provide strong empirical evidence of the redundancy in Transformers, and efficiency gains can be achieved via architecture design or system-level optimization, they usually identify removable components in trained models, leading to limited guidance on how architectural resources should be allocated at a priori, or why particular model configurations are effective in theory. 

To tackle these problems, in this work, we qualitatively study the topic of parameter efficiency (in Transformers), which is in fact two sides of the same coin: 
(i) With fixed resources/budgets of total parameters, how to configure the model to enhance performance? 
(ii) To maintain performance, can we and how to reduce resources/budgets of total parameters? 
Specifically, this parameter-performance trade-off in multi-head attention Transformers are regarding the allocation of attention heads and head dimensions. 
Particularly in this work, we consider the following two problems: 
(i) Given fixed model dimensions, how to allocate the number of heads and head dimensions for better model performance?  (ii) Without degrading model performance, can we reduce head dimensions with a fixed number of heads?  

To this end, for problem (i), we derive an approximation error estimate for information extraction tasks, 
which implies an intrinsic trade-off between the number of heads and head dimensions aiming at enhancing model performance with the fixed total parameter budget (i.e. fixed model dimension) in Section \ref{section: performance}. 
For problem (ii), we further theoretically prove a  saturation pattern in softmax-based attention, showing that continuously increasing head dimensions (with the fixed number of heads) leads to diminishing returns in terms of model capacity, especially for long input sequences (Section \ref{section: budget}). Together, these \emph{theoretical results} derived for the \emph{fundamental attention mechanism} provide principled guidance for efficient parameter allocation in Transformers, and are also consistently validated through extensive numerical experiments. 

Our main contributions are summarized as follows:
\begin{itemize}
    \item With the fixed model dimension as total parameter budget, we theoretically characterize the trade-off between the number of heads and head dimensions to improve the model performance on information extraction tasks (Theorem \ref{theorem: tradeoff}).
    
    \item With maintained performance, we mathematically prove a saturation pattern of softmax-based attention (Theorem \ref{theorem: softmax} and Corollary \ref{corollary: softmax}), showing diminishing returns in model expressiveness when continuously increasing head dimensions (the number of heads fixed). 
    
    \item Our theory is consistently verified through extensive experiments across multiple tasks ($n$-gram, next-token prediction and masked language modeling) and different models (customized  Transformers, \textsc{LLaMA}-3-8B and \textsc{LLaMA}-3.2-1B), yielding principled and data-agnostic parameter allocation strategies (Corollary~\ref{corollary: tradeoff allocation} and Algorithm~\ref{algorithm: reduction}).
\end{itemize}


\paragraph{Notations.} For $k\in\mathbb{N}_+$, let $[k]=\left\{1,2,...,k\right\}$. For a vector $\boldsymbol{v}$ and $1\leq p\leq\infty$, we denote the $\ell_p$-norm of $\boldsymbol{v}$ by $\|\boldsymbol{v}\|_{\ell_p}$. For a matrix $\boldsymbol{A}=(a_{ij})$, we denote the spectral norm of $\boldsymbol{A}$ by $\|\boldsymbol{A}\|_2$, and $\|\boldsymbol{A}\|_{\infty}$ denotes the maximum norm, i.e., the maximum absolute entry of $\boldsymbol{A}$; $\|\boldsymbol{A}\|_{2,\infty}$ is defined by first computing the $\ell_2$ norm of each column of $\boldsymbol{A}$, and then taking the maximum among them. For an event $S$, we define the indicator $\mathbb{I}_{\{S\}}$, which equals to $1$ when $S$ occurs and $0$ otherwise. We use the standard big-O notation $\mathcal{O}$ to hide absolute positive constants. For a matrix $\boldsymbol{X}$, we denote the $a$-th to $b$-th columns by $\boldsymbol{X}_{a:b}$.

\section{Preliminaries}

In this section, we introduce the Transformer architecture by formulating each single layer. 
The Transformer maps input sequences to output sequences in the following order. 

\paragraph{Input sequence.} We assume that the input sequence has the length $L$, and each token $\boldsymbol{x}_t\ (t\in[L])$ is in $\mathbb{R}^d$, hence the input sequence is $\boldsymbol{X}=(\boldsymbol{x}_t)_{t\in[L]}\in\mathbb{R}^{d\times L}$. 

\paragraph{Embedding layer.} For each token $\boldsymbol{x}_t$, we transform it into a $D$-dimensional vector through a linear mapping $\boldsymbol{x}_t^{(0)}=\boldsymbol{W}_E\boldsymbol{x}_t+\boldsymbol{b}_E$ ($\boldsymbol{W}_E\in\mathbb{R}^{D\times d}$, $\boldsymbol{b}_E\in \mathbb{R}^D$). We denote the embedded input sequence by  $\boldsymbol{X}^{(0)}\in\mathbb{R}^{D\times L}$.

\paragraph{Transformer layer.} 
The embedding $\boldsymbol{X}^{(0)}$ are then passed through $N$ Transformer layers, each composed of a multi-head self-attention (MHSA) and a feed forward network (FFN) layer: 
\begin{align*}
    \boldsymbol{X}^{(n-\frac{1}{2})}&=\boldsymbol{X}^{(n-1)}+\mathrm{MHSA}^{(n)}(\boldsymbol{X}^{(n-1)}),~~ n\in[N], \\
    \boldsymbol{X}^{(n)}&=\boldsymbol{X}^{(n-\frac{1}{2})}+\mathrm{FFN}^{(n)}(\boldsymbol{X}^{(n-\frac{1}{2})}),~~ n\in[N],  
\end{align*}
where 
\begin{align*}
  \mathrm{MHSA}^{(n)}(\boldsymbol{X}^{(n-1)})&=\boldsymbol{W}_O^{(n)}\cdot\mathrm{Concat}((\mathrm{Attn}^{(n,h)}(\boldsymbol{X}^{(n-1)}))_{h=1}^H),\\
  \mathrm{Attn}^{(n,h)}(\boldsymbol{X}^{(n-1)})
    &=\boldsymbol{W}_V^{(n,h)} \boldsymbol{X}^{(n-1)}
  \cdot
  \mathrm{softmax}(\langle\boldsymbol{W}_Q^{(n,h)}\boldsymbol{X}^{(n-1)},\boldsymbol{W}_K^{(n,h)}\boldsymbol{X}^{(n-1)}\rangle+\boldsymbol{R}^{(n,h)}),
\end{align*}
where $\boldsymbol{W}_K^{(n,h)}, \boldsymbol{W}_Q^{(n,h)},\boldsymbol{W}_V^{(n,h)}\in\mathbb{R}^{d_h^{(n)}\times D}$, and $\boldsymbol{W}_O^{(n)}\in\mathbb{R}^{D\times D}$ are learnable parameters (weight matrices) corresponding to the key, query, value and output matrix of the $h$-th head at the $n$-th layer, and $\big<\boldsymbol{A},\boldsymbol{B}\big>:=\boldsymbol{A}^\top\boldsymbol{B}$. The activation function $\mathrm{softmax}(\cdot)$ is performed in a column-wise sense, and  $\mathrm{Concat}(\cdot)$ acts vertically to concatenate $H$ matrices; 
$\boldsymbol{R}^{(n,h)}$ denotes the relative positional encoding (RPE) matrix (\cite{vaswani2017attention, devlin2019bert, shaw2018self, dai2019transformerxl, raffel2019t5, su2021roformer, press2022trainshorttestlong, huang2018music, yang2019xlnet}), which satisfies $\boldsymbol{R}_{ij}^{(n,h)}=-\infty$ if $i<
j$. 
For instance, in the Alibi RPE ~(\cite{press2022trainshorttestlong}), we have  $\boldsymbol{R}_{ij}^{(n,h)}=\phi(i-j;p^{(n,h)})$, where $p^{(n,h)}$ collects learnable parameters and 
\begin{equation*}
\phi(z; p)
=
\begin{cases}
-p\cdot z, & z \geq 0,\\
-\infty, & \mathrm{otherwise}.
\end{cases}
\end{equation*}


\paragraph{Feed forward network.} The feed forward network (FFN) is applied token-wisely, which maps each column of the internal sequence $\boldsymbol{X}^{(n-\frac{1}{2})}$ from $\mathbb{R}^D$ to $\mathbb{R}^D$ in the first $N-1$ layers, and from $\mathbb{R}^D$ back to $\mathbb{R}^d$ in the $N$-th layer. According to \cite{cybenko1989approximation, hornik1989multilayer, e2019apriori, lin2018resnet, li2020curse, li2019dynamical}, the FFN has universal approximation property as the number of parameters increases.

\section{Improving Performance with Fixed Parameter Budget}\label{section: performance}

In this section, we investigate problem (i) regarding parameter allocation in multi-head attention, aiming to determine the optimal number of heads and head dimensions given a fixed total parameter budget (i.e. a fixed model/embedding dimension) that minimizes the learning error. 
We present our approximation results in Section~\ref{section: tradeoff}, and validate them empirically in Section~\ref{section: exp1}.

\subsection{Parameter Trade-offs in Information Extraction}\label{section: tradeoff}


We study the information extraction task, formulated as learning linear combinations of input tokens. Many practical Transformer behaviors can be traced to such information extraction mechanisms. For example, sparse memory (\cite{wang2024understandingexpressivepowermechanisms}) extends n-gram-style pattern extraction, while induction heads implement copy-like operations underlying in-context learning and certain reasoning behaviors (\cite{olsson2022context}). 

Our analysis also extends to nonlinear sequence modeling. From an operator-learning perspective, Transformers approximate causal mappings over sequences, where information extraction corresponds to the linear regime and more complex dependencies arise from higher-order interactions. Such continuous and time-homogeneous nonlinear systems can be modeled by Volterra series, which have recently re-emerged in time-series modeling (\cite{huangrethinking}), signal processing (\cite{shiki2017application}), and quantitative finance (\cite{el2019characteristic}).

We first present results for the information extraction task, formulated as follows. 

\paragraph{Linear sequence modeling.} Assume that $\boldsymbol{x}_t$ are be random vectors defined on a common probability space. They are not
required to be independent, identically distributed, isotropic, or sub-Gaussian. Assume only that their $\ell_2$-norms are uniformly upper bounded by $B$. Consider a sequence-to-sequence modeling framework widely-adopted in prior theoretical studies (\cite{JMLR:v23:21-0368,li2020curse,wang2024stablessmalleviatingcursememory,pmlr-v139-jiang21d,wang2023inverse}):\footnote{This (linear) sequence modeling framework has been shown to yield important and profound insights into the interplay between model approximation property and data memory structure for multiple sequence models, such as SSMs (\cite{JMLR:v23:21-0368,li2020curse,wang2024stablessmalleviatingcursememory,wang2023inverse}), temporal CNNs (\cite{pmlr-v139-jiang21d}), and Transformers (\cite{jiang2024approximation}).} 
$\boldsymbol{y}_t=\mathrm{H}_t(\boldsymbol{X})=\sum_{i=0}^{L}\boldsymbol{\rho}_i\boldsymbol{x}_{t-i}$ (with a zero padding on $\boldsymbol{x}_{t}$), i.e., the output sequences are convolutions of input sequences with the target ``kernel'' $\boldsymbol{\rho}_i \in \mathbb{R}^{d \times d}$. 
For information extraction tasks, the target kernel $\boldsymbol{\rho}_i$ can be viewed as the extractor. 

Under this setting, the hypothesis space consists of single-layer Transformers with a fixed total parameter budget (i.e. the fixed embedding/model dimension) $D$. 
Let $\mathcal{T}^D$ denote all single-layer Transformers with a fixed embedding dimension $D$, then the approximation problem is defined as 
$$\mathcal{E}_D(\boldsymbol{X})=\inf_{\mathrm{T}\in\mathcal{T}^D}\|\mathrm{H}_t(\boldsymbol{X})-\mathrm{T}(\boldsymbol{X})\|_{2,\infty}.$$
In this formulation, with a fixed budget $D$, parameters (i.e. the number of heads and head dimensions) must be allocated judiciously to learn the target in the form of linear combinations of the most informative tokens. 
We establish the following approximation error estimate. 

\begin{theorem}\label{theorem: tradeoff}
    Suppose there are $H_m$ heads each with the dimension $d_m$ ($m\in [M]$), and the model dimension $D=\sum_{m=1}^M H_m\cdot d_m$ is fixed. To approximate the linear target $\mathrm{H}_t(\boldsymbol{X})=\sum_{i=0}^{L}\boldsymbol{\rho}_i\boldsymbol{x}_{t-i}$, for any $\delta \in (0,1)$, with the probability at least $1-\delta$, we have
    \begin{equation}\label{equation: E(D)}
    \begin{aligned}
    \mathcal{E}_D(X)\le& \quad B\sum_{m=1}^{M}\|\rho_m\|_2
    \left[\mathbb{I}_{\{d_m<d\}}\left(\sqrt{1-\frac{d_m}{d}}\right)+\frac{1.3e^{0.02m}}{H_m}\right]+B\sum_{k=M+1}^{L}\|\rho_k\|_2\\
    &+B\sum_{m=1}^{M}\|\rho_m\|_2 \cdot C\sqrt{\frac{\log(2M/\delta)}{d}}.
    \end{aligned}
    \end{equation}
    \end{theorem}


    We provide the detailed proof of Theorem \ref{theorem: tradeoff} in Appendix~\ref{appendix: proof1}, and briefly summarize the main construction ideas as follows: 
    \begin{enumerate}
        \item Since the embedding dimension is fixed, we first compress the information contained in each token, which yields the first term in the estimate (Eq. (\ref{equation: E(D)})). 
        \item We then employ $M$ groups of heads, where each group is responsible for extracting a single token ahead. The approximation error decreases as more heads are employed, corresponding to the second term in the estimate. 
        \item Finally, since only the $M$ most informative tokens are extracted, which truncates the target at the time step $M$, we derive the third term in the estimate. 
        \item Without loss of generality, $\boldsymbol{\rho}_i$ can be reordered so that $||\boldsymbol{\rho}_i||_2$ decreases with $i$, minimizing the third term.
    \end{enumerate}
    According to Theorem \ref{theorem: tradeoff}, 
    on one hand, setting a large head dimension is beneficial for preserving token information, reducing the first error term in equation (\ref{equation: E(D)}); 
    on the other hand, increasing the number of heads is also favorable for approximating the extraction function, reducing the second error term. 
    However, since the total budget/number of parameters is fixed, one cannot increase the number of heads and head dimensions simultaneously beyond certain constraints, which inevitably leads to a trade-off, see Appendix~\ref{appendix: tradeoff} for illustration. 

Here, we formulate the information extraction task as learning to construct linear combinations of input tokens, typically represented as a convolutional form. For a single Transformer layer, \cite{jiang2025numericalinvestigationsequencemodeling} also identifies a trade-off phenomenon in this type of tasks, as illustrated in their experiments (see Figure~\ref{exp: linear tradeoff} in Appendix~\ref{appendix: experiment}): 
The trade-off between the number of attention heads and learning errors emerges in different sequence modeling tasks, with its manifestation depending on the type and strength of underlying memory structures captured by the convolution kernel $\boldsymbol{\rho}$. Their results demonstrate that these parameter-performance trade-offs consistently arise across a variety of memory structures (e.g., exponentially/polynomially decaying kernels, delta/Airy function kernels). However, in certain cases such as exponentially and polynomially decaying kernels, these trade-offs may vanish when the memory strength varies. Theorem \ref{theorem: tradeoff} also provides a theoretical explanation for this phenomenon.

Based on Theorem \ref{theorem: tradeoff}, we consequently have the following parameter allocation strategy.  

\begin{corollary}[Principled parameter allocation]\label{corollary: tradeoff allocation}
    Under the same conditions of Theorem \ref{theorem: tradeoff}, the optimal allocation of parameters can be achieved by solving the following optimization problem
\begin{equation}\label{equation: tradeoff opt}
\begin{aligned}
    \min_{M,H_m,d_m}&\quad\sum_{m=1}^M\|\boldsymbol{\rho}_m\|_2\left(\mathbb{I}_{\left\{d_m\leq d\right\}}\sqrt{1-\frac{d_m}{d}}+\frac{1.3\  e^{0.02m}}{H_m}\right)+\sum_{k=M+1}^{L}\|\boldsymbol{\rho}_k\|_2,\\
    \mathrm{s.t.}&\quad \sum_{m=1}^M H_m\cdot d_m=D.
\end{aligned}
\end{equation}
\end{corollary}

Corollary \ref{corollary: tradeoff allocation} is a direct conclusion of Theorem \ref{theorem: tradeoff}. 
Next, we provide some specific tasks for illustrations of Corollary \ref{corollary: tradeoff allocation}.  

\paragraph{Example I: N-gram.} The $n$-gram task in language modeling involves predicting the next token in a sequence using the previous $n$ tokens as contexts (\cite{katz1987estimation, kneser1995improved, Chelba2017Ngram, Buck2014CommonCrawl, wang2024understandingexpressivepowermechanisms}). This task aims to capture local statistical dependencies in texts and serves as a basic benchmark for language modeling. We take a simple $4$-gram task as an example, where 
\begin{equation}\label{equation: 4 gram}
    \boldsymbol{y}_t = \boldsymbol{x}_{t-1}+\boldsymbol{x}_{t-2}+\boldsymbol{x}_{t-3}+\boldsymbol{x}_{t-4}.
\end{equation}

In this case, $\boldsymbol{\rho}_i=\mathbb{I}_{\left\{1\leq i\leq 4\right\}}\boldsymbol{I}_{d\times d}$. Let $d=8, D=256$ and $B=1$, then Eq. (\ref{equation: tradeoff opt}) becomes
\begin{equation*}
\begin{aligned}
    \min_{M,H_m,d_m}&\quad\sum_{m=1}^M\mathbb{I}_{\left\{d_m\leq 8\right\}}\sqrt{1-\frac{d_m}{8}}+ \sum_{m=1}^{M}\frac{1.3\  e^{0.02m}}{H_m}+(4-M),
   \\
    \mathrm{s.t.}&\quad \sum_{m=1}^M H_m\cdot d_m=256, \quad M\leq 4. 
\end{aligned}
\end{equation*}
We find numerically by search that the optimal solution is attained at $M=4,\ d_m=H_m=8\ (m=1,2,3,4)$. See further experimental validations in Figure~\ref{exp: 4-gram tradeoff}.

\paragraph{Example II: Induction head.} The induction head (IH) is a mechanism in Transformers that learns to copy patterns by attending from a repeated token to its earlier occurrence, enabling sequence continuation beyond training contexts~(\cite{elhage2021mathematical}). Implementing an IH requires two Transformer layers~(\cite{sanford2024onelayertransformersfailsolve,sanford2024transformersparallelcomputationlogarithmic,rajaraman2024transformersmarkovdataconstant,wang2025transformersrichapproximationdynamics}). In this construction, the first layer plays a role similar to an $n$-gram task, extracting linear combinations of tokens. This allows us to derive an allocation strategy, with details provided in Appendix~\ref{appendix: ih}.

\paragraph{Nonlinear sequence modeling.} 
Theorem~\ref{theorem: tradeoff} can be further extended to nonlinear settings.
Specifically, we consider a continuous, time-homogeneous system that admits the decomposition in 
\cite{wang2024stablessmalleviatingcursememory,1456575}. 
The corresponding results and proofs are deferred to Appendix~\ref{appendix: volterra}.

\subsection{Empirical Results}\label{section: exp1}

In this section, we present experimental results that validate Theorem~\ref{theorem: tradeoff} or Corollary~\ref{corollary: tradeoff allocation}. 
It is numerically shown that the trade-off between the number of attention heads and head dimensions exists to enhance model performance under a fixed total parameter budget. 


% ...

\begin{figure*}[htbp]
    \centering
    
    \begin{subfigure}[b]{0.325\linewidth}
        \centering
        \includegraphics[width=\linewidth]{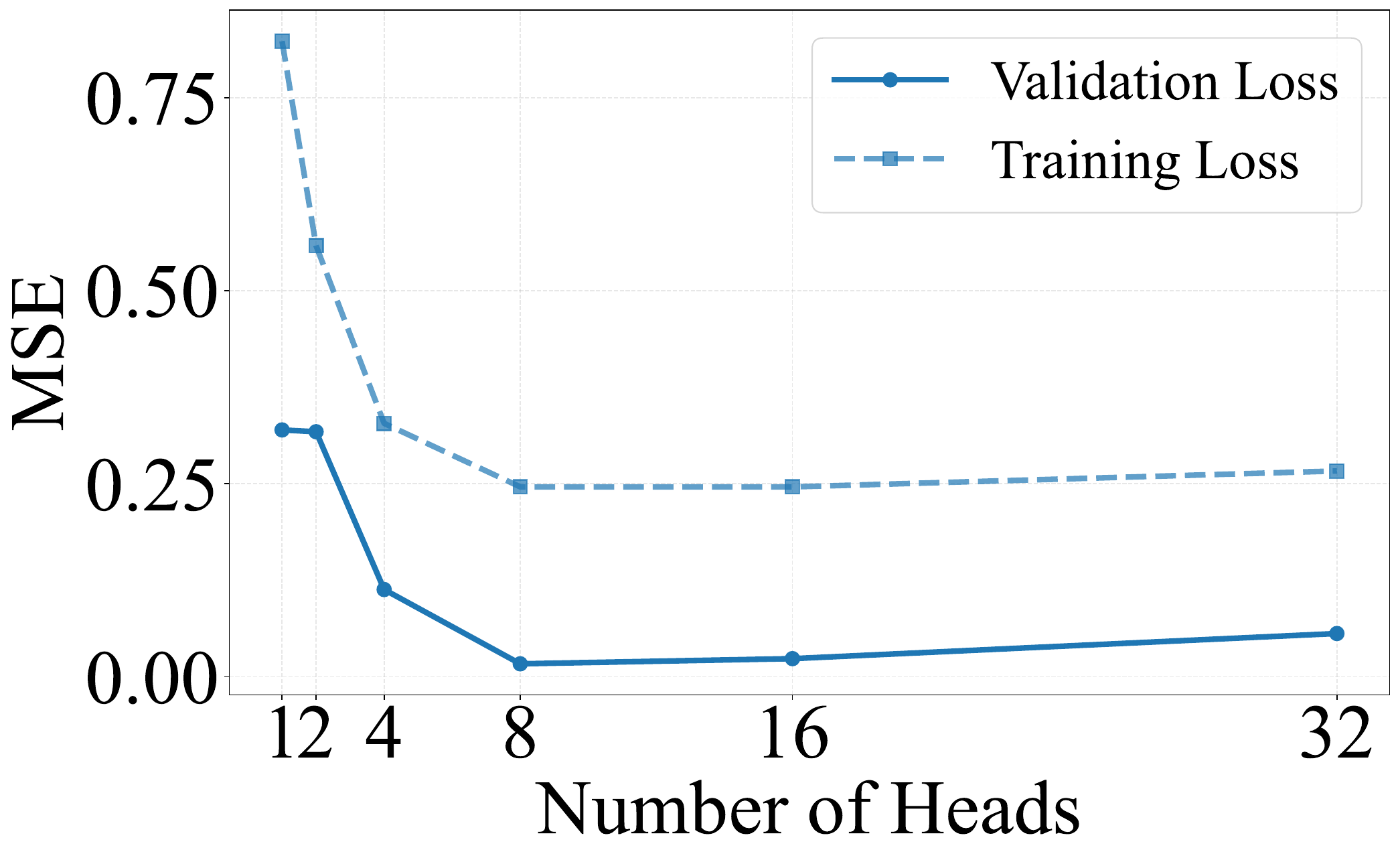}
        \caption{$N$-gram task}
        \label{exp: 4-gram tradeoff}
    \end{subfigure}
    \hfill
    \begin{subfigure}[b]{0.325\linewidth}
        \centering
        \includegraphics[width=\linewidth]{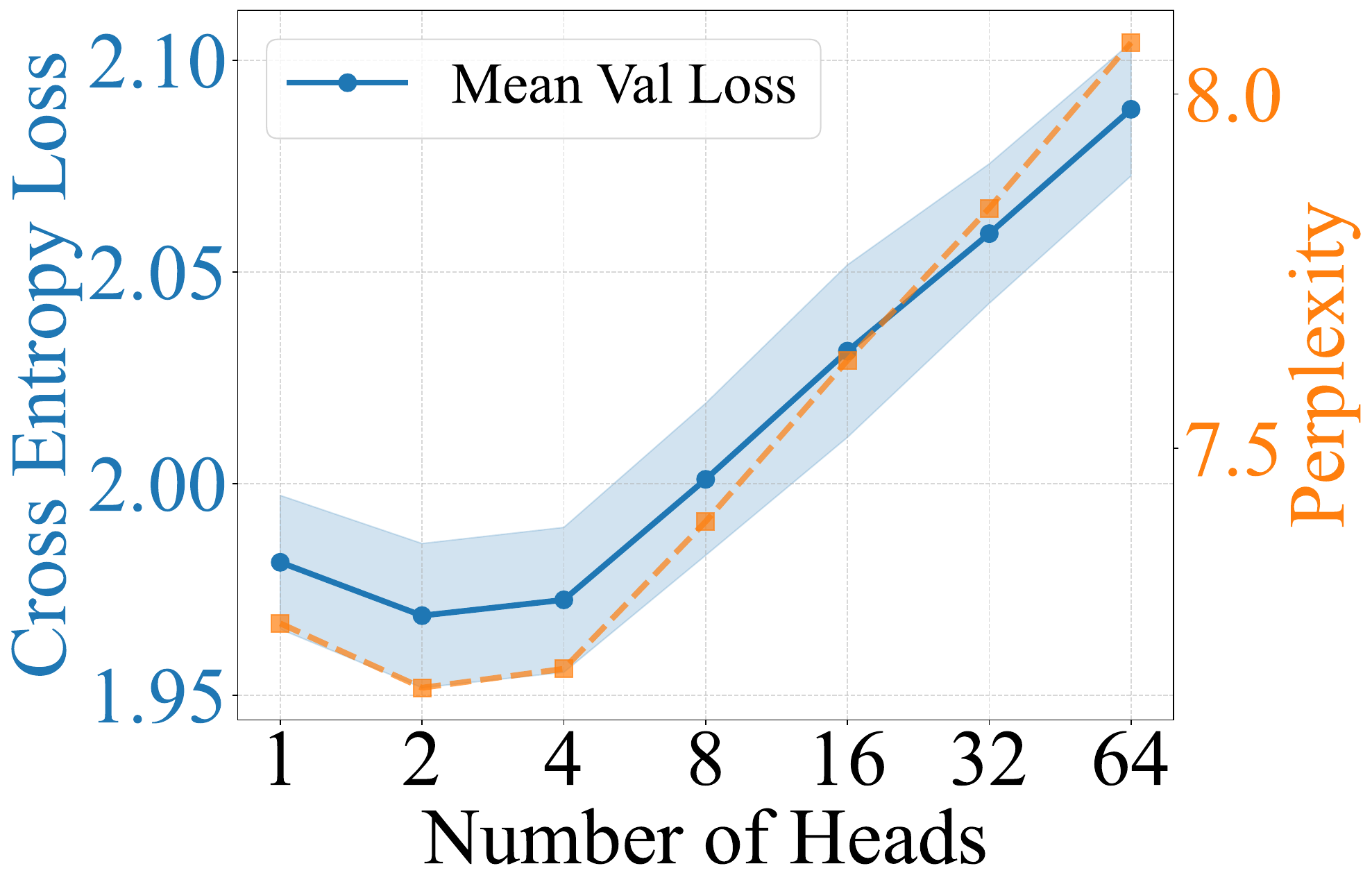}
        \caption{Next-token prediction}
        \label{exp: IR}
    \end{subfigure}
    \hfill
    \begin{subfigure}[b]{0.325\linewidth}
        \centering
        \includegraphics[width=\linewidth]{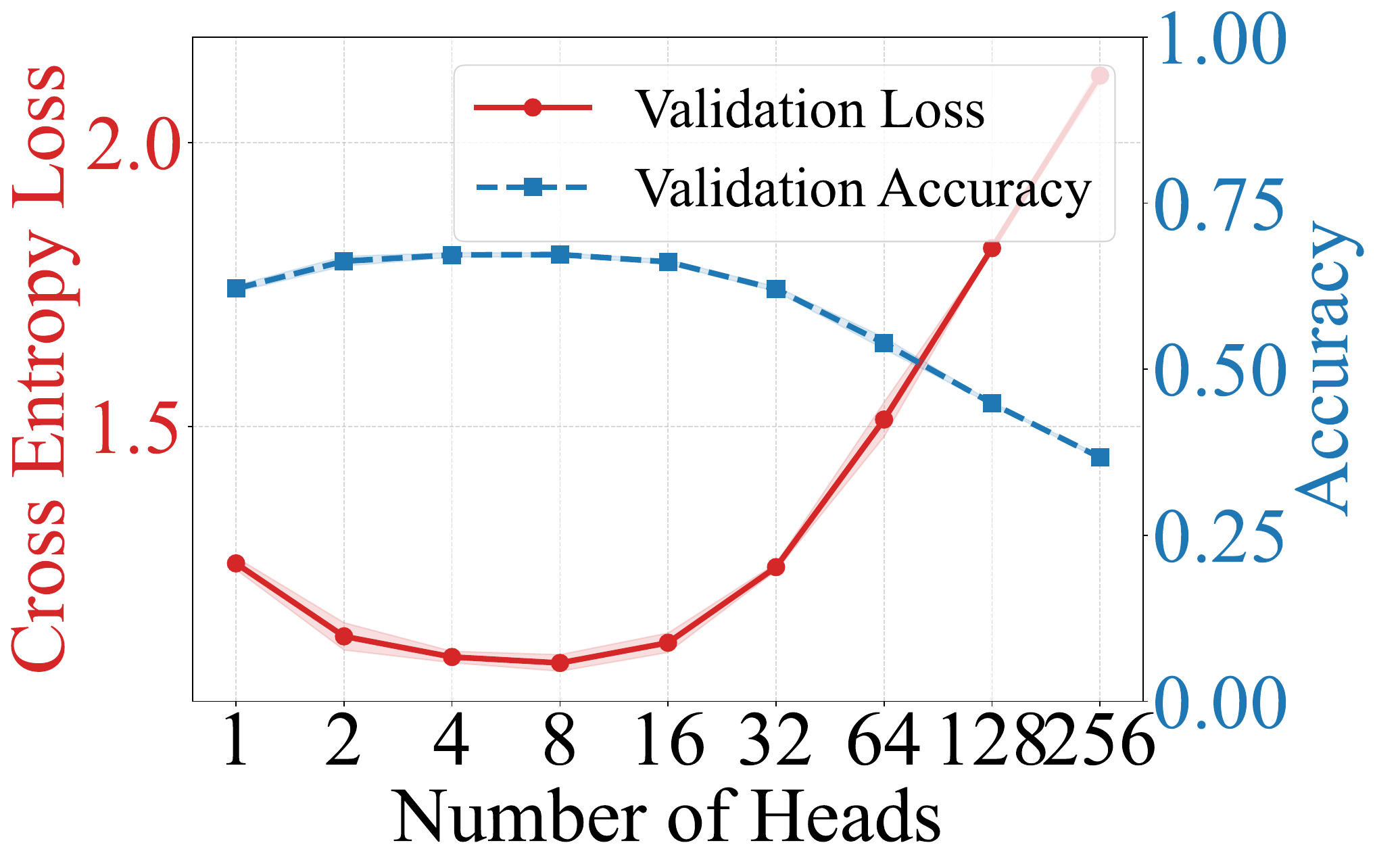}
        \caption{Masked language modeling}
        \label{exp: MLM}
    \end{subfigure}
    \caption{Trade-off to enhance model performance with a fixed total parameter budget.\protect\footnotemark[1]}
    \label{fig: all-plots}
\end{figure*}

\paragraph{Verification of Corollary~\ref{corollary: tradeoff allocation}.}
We empirically validate the parameter allocation strategy on the above $4$-gram task (Eq.~(\ref{equation: 4 gram})). A single-layer Transformer encoder with sinusoidal positional encoding~(\cite{vaswani2017attention}) is trained on a synthetic dataset (white noises) to learn the target Eq.~(\ref{equation: 4 gram}). With a fixed embedding dimension $D=256$, we conduct experiments under multiple learning rates and random seeds, and report the results corresponding to the best-performed configuration as an estimate of the approximation error. As shown in Figure~\ref{exp: 4-gram tradeoff}, the optimal allocation occurs at $M=4$ with $d_m=H_m=8$ $(m=1,2,3,4)$, 
which is consistent with the above theoretical prediction and confirms the proposed strategy (Eq.~(\ref{equation: tradeoff opt})).

\footnotetext[1]{For other values of $D$ in Figure \ref{fig: all-plots}(b) and (c), please refer to Appendix \ref{appendix: further task} (see Figure \ref{exp: ntp256}, \ref{exp: ntp512}, \ref{exp: mlm128} and \ref{exp: mlm512}).}

\paragraph{More trade-off examples.}

In Figure~\ref{exp: IR} and~\ref{exp: MLM}, we also investigate the trade-off phenomenon under a fixed total parameter budget.
In Figure~\ref{exp: IR}, a three-layer causal Transformer decoder is trained on the Tiny Shakespeare corpus for the next-token prediction task. 
Fix the bottleneck dimension as $D = 128$ , we vary the number of attention heads and evaluate model performance using both cross entropy loss and perplexity (PPL), with results averaged over multiple random seeds.
In Figure~\ref{exp: MLM}, we consider a masked language modeling (MLM) task by training a three-layer character-level Transformer encoder (with full self-attention) also on the Tiny Shakespeare corpus.
Following the standard MLM protocol \cite{devlin2019bert}, 15\% of input tokens are randomly masked, and the model is trained to reconstruct the masked tokens while ignoring unmasked positions during optimization.
Fix the model dimension as $D = 256$, we report both cross-entropy loss and masked token prediction accuracy, with all results averaged over three independent runs.
Across both experiments, we consistently observe a clear trade-off phenomenon, demonstrating that appropriate architectural choices can improve model performance under a fixed total parameter budget. We additionally include code, math, and logic-intensive tasks on a $330M$ model to broaden the scope in Appendix \ref{appendix: further task} Figure \ref{exp: math}.

\paragraph{Practical guideline.} Based on these observations, we suggest that practitioners should start with relatively smaller $H$ (thus larger $d_H$), and then perform a local sweep over a small range of $H$ to identify the optimal tradeoff point. In particular, overly large $H$ should be avoided, as excessively small $d_H$ can hurt performance even in simple tasks.

\section{Reducing Parameters with Maintained Performance}\label{section: budget}

In this section, we turn to problem (ii), aiming to reduce the required total parameters  by decreasing head dimensions without significantly degrading model performance. 
This is provably guaranteed by the saturation pattern of softmax-based attention, meaning that continuously increasing head dimensions can lead to diminishing returns in learning errors, particularly for long sequences. 
This saturation pattern is first mathematically proved, implying potential model efficiency guidance with reduced parameters (head dimensions) while maintaining ``lossless'' model capacity. 



\subsection{Softmax Saturation and Parameter Reduction}

The theory in this section is motivated by understanding modern architectures such as multi-head latent attention (MLA; \cite{liu2024deepseek}), as well as the phenomenon of rank collapse in attention. While several prior works have pointed out the low-rank nature of attention, the theoretical understanding remains incomplete. Our results contribute to this line of research by quantifying a redundancy-related estimate particularly when the sequence length is sufficiently large.

We first introduce the notation. Let 
\begin{align*}
    \mathrm{Logits}_{d_H}(\boldsymbol{W}_Q,\boldsymbol{W}_K,\boldsymbol{X})\quad&:= \langle\boldsymbol{W}_Q\boldsymbol{X},\boldsymbol{W}_K\boldsymbol{X}\rangle+\boldsymbol{R} \in \mathbb{R}^{L\times L}, 
\end{align*}
with $\boldsymbol{X}\in\mathbb{R}^{D\times L}$, $\boldsymbol{W}_K, \boldsymbol{W}_Q\in\mathbb{R}^{d_H\times D}$, and $\boldsymbol{R} \in\mathbb{R}^{L\times L}$. For $l\in [L]$, let $\boldsymbol{e}_l$ be the vector with the $l$-th element as $1$ and all other elements as $0$.
For each query position $l\in[L]$, let
\[
    \mathcal S_l:=\{j\in[L]:R_{lj}>-\infty\},
    \qquad s_l:=|\mathcal S_l|,
\]
be the set and number of unmasked key positions. Define the $l$-th
query logit vector and attention vector, restricted to their active
coordinates, by
\[
    z_l:=\bigl(e_l^\top\mathrm{Logits}_{d_H}\bigr)_{\mathcal S_l},
    \qquad
    a_l:=\operatorname{softmax}(z_l),
\]
and let
\[
    J_l:=\frac{\partial a_l}{\partial z_l}
        =\operatorname{diag}(a_l)-a_la_l^\top.
\]
Thus $s_l=L$ in the unmasked case, whereas $s_l=l$ under the standard
causal mask.
Then, we have the following estimate on the column-wise  Jacobian of softmax-based attention.


\begin{theorem}[Softmax sensitivity under bounded active logits]
\label{theorem: softmax}
Assume
$
    \|W_Q\|_2\le M_Q,
    \,
    \|W_K\|_2\le M_K,
    \,
    \|x_t\|_2\le B\, (t\in[L]),
$
and that the finite positional-bias entries satisfy
$
    |R_{lj}|\le R_0\,(j\in\mathcal S_l,\ l\in[L]).
$
Let
$
    \Delta:=M_QM_KB^2+R_0.
$
Then, for every $l\in[L]$,
$$
    \|J_l\|_2
    \le
    \frac{2e^{2\Delta}}{s_l}.
$$
Consequently,
$$
    \|J_l\|_2=\mathcal O(L^{-1})
$$
for unmasked attention, while under the standard causal mask,
$
    \|J_l\|_2=\mathcal O(l^{-1}).
$
The hidden constants depend only on $M_Q,M_K,B$, and $R_0$.
\end{theorem}


According to Theorem \ref{theorem: softmax}, the ``slope'' of softmax activations is quite flat, particularly for long sequences. 
This further implies the following corollary, which characterizes the approximation error when ``compressing'' attention score matrices with reduced head dimensions. 


\begin{corollary}[Head-dimension reduction]
\label{corollary: softmax}
Under the assumptions of Theorem~\ref{theorem: softmax}, let
$
    M:=W_Q^\top W_K,
$
and let $M_{d_h}$ be its rank-$d_h$ truncated SVD. Choose
$\hat{W}_Q,\hat{W}_K\in\mathbb R^{d_h\times D}$ such that
$
    \hat{W}_Q^\top\hat{W}_K=M_{d_h}.
$
Let $\hat A$ denote the attention-score matrix obtained by replacing
$W_Q,W_K$ with $\hat{W}_Q,\hat{W}_K$ while keeping the same
positional bias and mask. Then for every query $l\in[L]$,
\[
    \left\|
    e_l^\top\bigl(
        \mathrm{AttnScore}_{d_H}-\hat A
    \bigr)
    \right\|_2
    \le
    \frac{2e^{2\Delta}}{s_l}
    \|X\|_2^2
    \sigma_{d_h+1}(M).
\]
Hence the per-query approximation error is
\[
    \mathcal O\!\left(
        \frac{\|X\|_2^2\sigma_{d_h+1}(M)}{L}
    \right)
\]
in the unmasked case and
\[
    \mathcal O\!\left(
        \frac{\|X\|_2^2\sigma_{d_h+1}(M)}{l}
    \right)
\]
under the standard causal mask.

In particular, for any fixed $\alpha\in(0,1)$, the late-query block
$\mathcal I_\alpha:=\{l:\,l\ge\lceil\alpha L\rceil\}$ satisfies
\begin{equation}\label{equation: teacher student error}
    \left(
    \sum_{l\in\mathcal I_\alpha}
    \left\|
    e_l^\top\bigl(
        \mathrm{AttnScore}_{d_H}-\widehat A
    \bigr)
    \right\|_2^2
    \right)^{1/2}
    \le
    \frac{2e^{2\Delta}}{\alpha\sqrt L}
    \|X\|_2^2
    \sigma_{d_h+1}(M).
\end{equation}
\end{corollary}

\begin{remark}[Causal masking and early queries]
For causal attention, the global Frobenius error over all queries does
not in general inherit an $L^{-1/2}$ rate from Theorem~\ref{theorem: softmax},
because
\[
    \sum_{l=1}^L l^{-2}=\mathcal O(1).
\]
The vanishing $L^{-1/2}$ Frobenius rate remains valid for any fixed
late-query fraction $l\ge\alpha L$. Early queries have much smaller
active support and, moreover, $\operatorname{rank}(J_l)\le l-1$.
\end{remark}

\paragraph{Extension to rotary positional embeddings.}
The preceding analysis uses an additive relative positional bias.
The same head-dimension reduction principle can also be extended to
rotary positional embeddings (RoPE), although the corresponding
low-rank approximation takes a slightly different form.
Let $\mathcal R_t\in\mathbb R^{d_H\times d_H}$ denote the RoPE
rotation at position $t$. Since $\mathcal R_t$ is orthogonal, the
relative rotation
\[
    \mathcal R_{l,j}:=\mathcal R_l^\top \mathcal R_j
\]
also satisfies
\[
    \mathcal R_{l,j}^\top\mathcal R_{l,j}=I,
    \qquad
    \|\mathcal R_{l,j}\|_2=1.
\]
Accordingly, the $(l,j)$-th attention logit can be written as
\[
    z_{lj}
    =
    x_l^\top
    W_Q^\top
    \mathcal R_{l,j}
    W_K
    x_j.
\]

Let $W_{Q,r}$ and $W_{K,r}$ denote the rank-$r$ truncated-SVD
approximations of $W_Q$ and $W_K$, respectively. Then, uniformly over
all relative positions,
\[
\begin{aligned}
\left\|
W_Q^\top\mathcal R_{l,j}W_K
-
W_{Q,r}^\top\mathcal R_{l,j}W_{K,r}
\right\|_2
\le
\|W_Q\|_2\,\sigma_{r+1}(W_K)
+
\sigma_{r+1}(W_Q)\,\|W_K\|_2.
\end{aligned}
\]
Thus, the approximation error caused by reducing the Q/K head
dimension remains controlled uniformly over the RoPE rotations.
Combining this estimate with the softmax sensitivity bound in
Theorem~\ref{theorem: softmax} yields the same inverse-support-size
saturation behavior. In particular, under causal masking, the
sensitivity scales as $\mathcal O(1/l)$ for the $l$-th query.
A formal statement and proof are provided in
Appendix~\ref{appendix: rope}.

The detailed proofs of Theorem \ref{theorem: softmax} and Corollary \ref{corollary: softmax} are provided in Appendix \ref{appendix: proof2}. 
Specifically, Corollary~\ref{corollary: softmax} is derived by applying an SVD-based low-rank approximation to original logits matrices, together with Theorem \ref{theorem: softmax}. According to Corollary~\ref{corollary: softmax}, for long sequences ($L\gg 1$), compressing original attention score matrices in large head dimensions with those in smaller head dimensions can lead to minor errors, potentially implying a principled strategy for additional parameter reduction with approximately maintained model capacity.

\paragraph{Principled strategy for parameter reduction.} 
Corollary~\ref{corollary: softmax} shows that a student head with reduced dimensions can approximate a teacher head with larger dimensions by incurring only small errors, particularly for long sequences. 
Therefore, a principled and practical strategy for additional parameter reduction can be motivated as Algorithm \ref{algorithm: reduction}. This procedure enables further model compression with enhanced parameter efficiency and lossless learning performance.

\begin{algorithm}[tb]
   \caption{Head Dimension Reduction Strategy}
   \label{algorithm: reduction}
\begin{algorithmic}
   \STATE {\bfseries Input:} Teacher parameters $\boldsymbol{W}_Q, \boldsymbol{W}_K \in \mathbb{R}^{d_H \times D}$, student dimension $d_h$
   \STATE {\bfseries Output:} Approximated student parameters $\hat{\boldsymbol{W}}_Q, \hat{\boldsymbol{W}}_K \in \mathbb{R}^{d_h \times D}$
   \STATE \textbf{1. SVD Initialization:} Initialize $\hat{\boldsymbol{W}}_Q, \hat{\boldsymbol{W}}_K$ via truncated SVD of $\boldsymbol{W}_Q^\top \boldsymbol{W}_K$ 
   \STATE \textbf{2. Refinement :} Further train $\hat{\boldsymbol{W}}_Q, \hat{\boldsymbol{W}}_K$ with other parameters fixed 
\end{algorithmic}
\end{algorithm}

\begin{remark}
The dimension-reduction strategy based on Corollary \ref{corollary: softmax} is complementary to existing low-rank techniques rather than a direct replacement. Building upon existing low-rank methods such as GQA (mainly used in \textsc{LLaMA} in Figure \ref{exp: low rank} and Figure \ref{exp: two stage}), our approach further reduces the parameter budget without significantly degrading model performance. 
Rather than designing new architectures or training compression-aware models from scratch, we consider the scenario where a large pretrained model is already available, and the goal is to compress it into a smaller model with reduced computation costs and lossless performance. This setting is particularly relevant for deployment on resource-constrained devices or for users with limited computation resources.
\end{remark}

\paragraph{Complexity analysis.}

We consider computation and memory costs arising from self-attention. 
By reducing $\boldsymbol{W}_Q, \boldsymbol{W}_K$ with $\hat{\boldsymbol{W}}_Q, \hat{\boldsymbol{W}}_K$, the computation complexity for linear projections $\hat{\boldsymbol{Q}}=\hat{\boldsymbol{W}}_Q\boldsymbol{X}$ and $\hat{\boldsymbol{K}}=\hat{\boldsymbol{W}}_K\boldsymbol{X}$ drops to $\mathcal{O}((d_h/d_H)^2LD^2)$, and $\mathcal{O}((d_h/d_H)L^2D)$ for the attention score $\langle \hat{\boldsymbol{Q}}, \hat{\boldsymbol{K}}\rangle$,  definitely saving the parameters budget for storing query/key matrices (\cite{vaswani2017attention, wang2020linformer}).



\subsection{Empirical Results}\label{section: budget exp}

In this section, we first numerically verify  Theorem~\ref{theorem: softmax} and Corollary~\ref{corollary: softmax}.
We then implement Algorithm~\ref{algorithm: reduction} on LLM-scale experiments (\textsc{LLaMA}-3-8B and \textsc{LLaMA}-3.2-1B), to empirically demonstrate that compressing head dimensions does not significantly degrade model performance for both pre-training and down-stream tasks. 
Finally, we validate the efficiency gains of the proposed algorithm. 

\paragraph{Verification of Theorem \ref{theorem: softmax}.} 

\begin{figure}[htbp]
    \centering
    \begin{minipage}[t]{0.32\linewidth} 
        \centering
        \includegraphics[width=\linewidth]{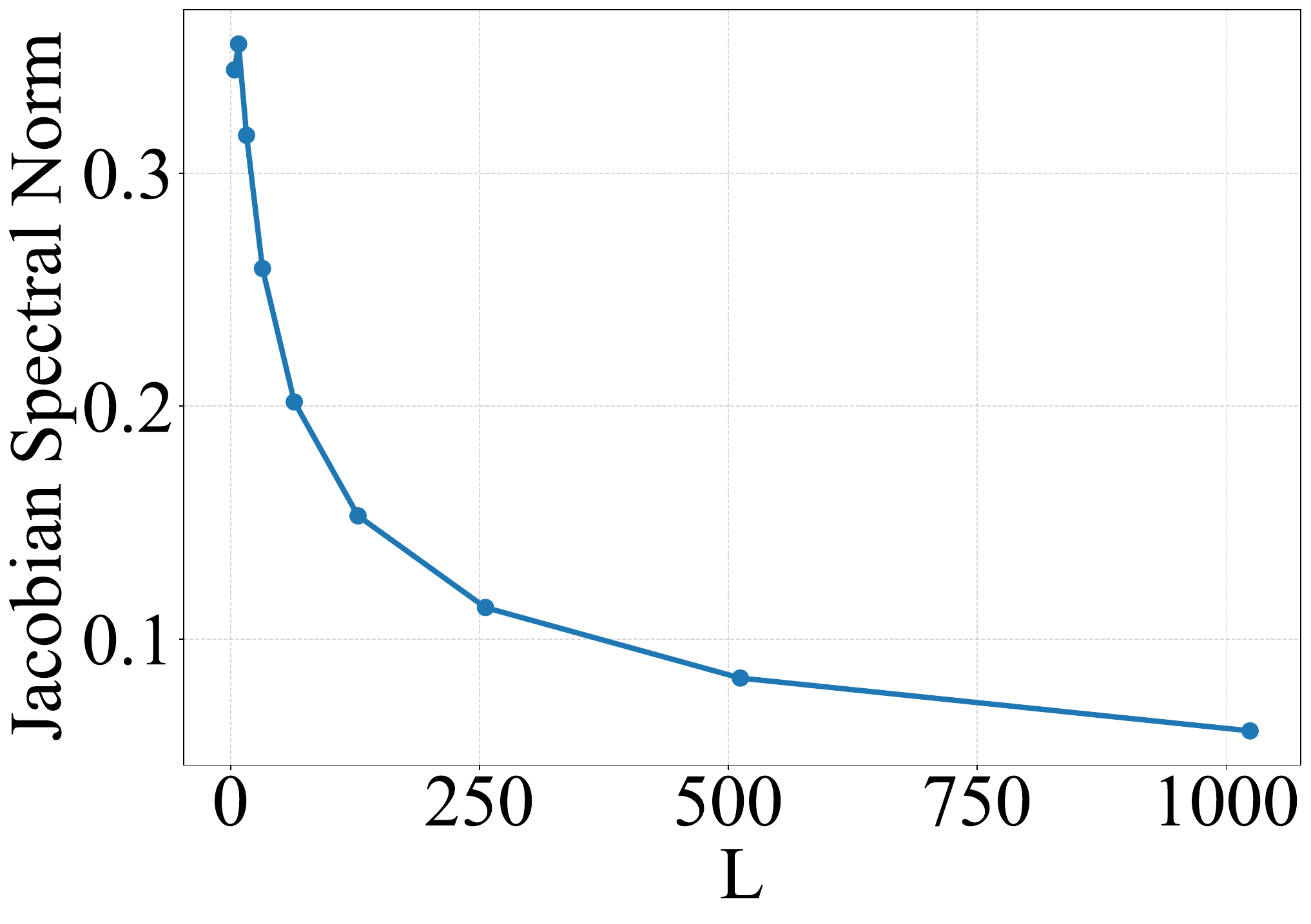}
        \caption{Numerical verification of Theorem \ref{theorem: softmax}: Softmax-based attention becomes flat (w.r.t. logits) for long sequences.}
        \label{exp: jacobian spectral}
    \end{minipage}
    \hspace{1mm}
    \begin{minipage}[t]{0.32\linewidth} 
        \centering
        \includegraphics[width=\linewidth]{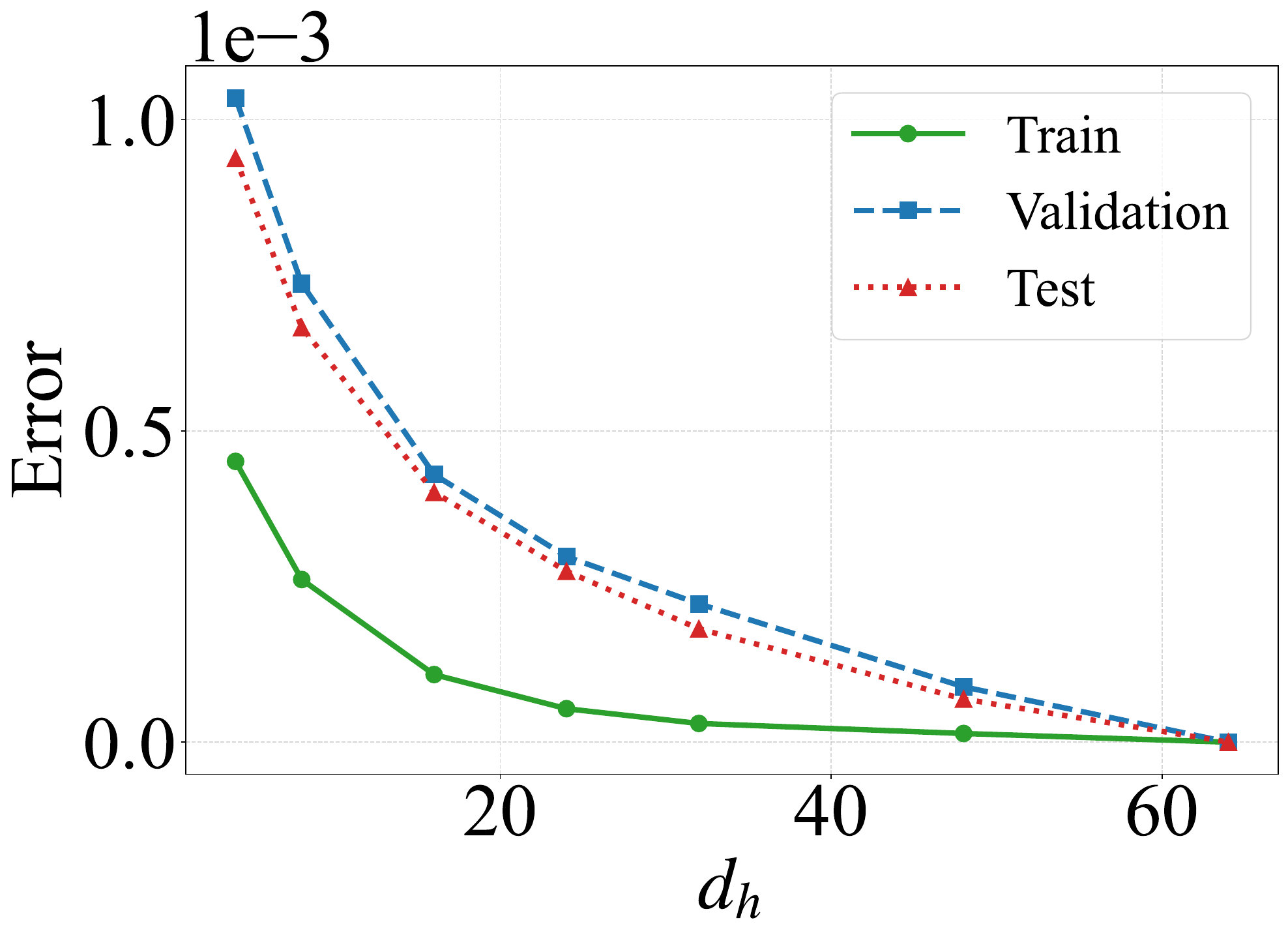}
        \caption{Numerical verification of Corollary \ref{corollary: softmax}: Students with reduced dimensions ($d_h$) can approximate teacher heads ($d_H=64>d_h$).}
        \label{exp: trained L scale}
    \end{minipage}
    \hspace{1mm}
    \begin{minipage}[t]{0.32\linewidth} 
        \centering
        \includegraphics[width=\linewidth]{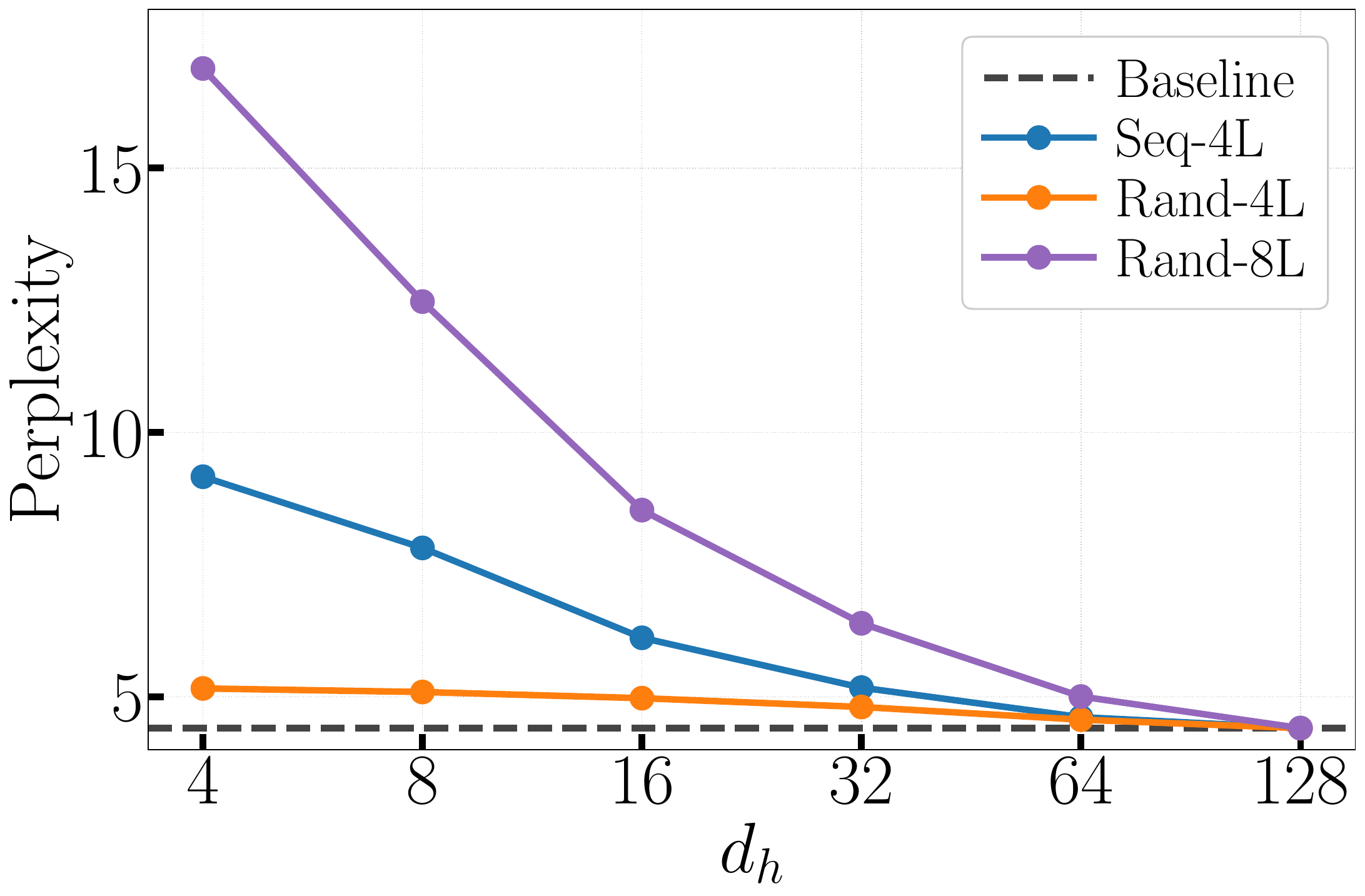}
        \caption{Reduced head dimensions in \textsc{LLaMA}-3-8B lead to saturated perplexity on WikiText-2.}
        \label{exp: low rank}
    \end{minipage}
\end{figure}

In Figure~\ref{exp: jacobian spectral}, we empirically validate Theorem~\ref{theorem: softmax} by analyzing how the spectral norm of the attention Jacobian varies with sequence lengths. Specifically, we study the TinyLlama-1.1B model 
and focus on Layer~0 and Head~0, using sequences from the WikiText-103 dataset. For sequence lengths $L \in \{4, 8, 16, 32, 64, 128, 256, 512, 1024\}$, we sample $10$ different sequences for each length.\footnotemark[2] For each sequence, we compute the attention Jacobian with respect to logits, calculate its spectral norm, and take average over the ten sequences.  

\paragraph{Verification of Corollary \ref{corollary: softmax}.} In Figure~\ref{exp: trained L scale}, we numerically validate Corollary \ref{corollary: softmax} by analyzing the error to fit attention heads of pre-trained Transformer language models with those of reduced head dimensions. Using the LiteLlama-460M
model as the teacher, we isolate a single teacher head (Layer~5, Head~0) and train a student head with reduced query-key dimensions $d_h \in \{4, 8, 16, 24, 32, 48, 64\}$ (input and rotary positional embeddings are retained). 
For $d_h < d_H=64$, the student query and key matrices are initialized via truncated SVD of the teacher’s. The training objective is to fit the teacher attention weights on a mixed dataset of synthetic and WikiText-103 sequences with the length $1024$.\footnotemark[1] We use the Adam optimizer with the learning rate $10^{-3}$ for 10,000 epochs, and the performance is evaluated by mean squared error (Eq. (\ref{equation: teacher student error})) over masked causal attention weights. 


The above experiments reflect that a single attention head can be effectively compressed via head dimension reduction for real-world language models and datasets. 
We further show that head dimension reduction applied to multi-head attention can largely maintain LLM's performance in terms of both pre-training perplexity and downstream accuracy. 

\footnotetext[2]{Each sequence is generated by randomly selecting contiguous spans from WikiText-103, which is tokenized to a fixed length $L$ (padded if necessary). For large $L$, padding ensures a consistent input size while maintaining semantically meaningful contents.}



\paragraph{\textsc{LLaMA}-3-8B: Saturated perplexity.} 
To further study the model reduction effect of pre-trained large Transformer models, we apply head-wise dimension reduction to the multi-head self-attention modules of \textsc{LLaMA}-3-8B (with an original head dimension $d_H=128$). Specifically, for selected layers, we parameterize reduced models with query/key weight matrices as low-rank decompositions (rank-$d_h<d_H$) of those in original \textsc{LLaMA}-3-8B, and retain all the other configurations 
These low-rank weights are initialized via truncated SVD of those in the original model, and we freeze all the other original parameters during training, to minimize the mean squared error (MSE) between the reconstructed low-rank weights and their original (full-rank) counterparts. The training is performed using the AdamW optimizer (with a learning rate $2\times 10^{-4}$) for 150 steps.  
We evaluate performance using perplexity (PPL) on WikiText-2 via a sliding-window protocol. Experiments cover multiple head dimensions $d_h \in \{4, 8, 32, 64, 128\}$ across different layers: \textit{Seq-4L} (four sequential layers), \textit{Rand-4L} (four random layers), and \textit{Rand-8L} (eight random layers). As shown in Figure~\ref{exp: low rank}, increasing head dimensions yields diminishing returns in perplexity, indicating that head dimensions can be reduced with approximately maintained model performance.



\begin{figure}[htbp]
    \centering
    \begin{minipage}[t]{0.475\linewidth}
        \centering
        \includegraphics[width=\linewidth]{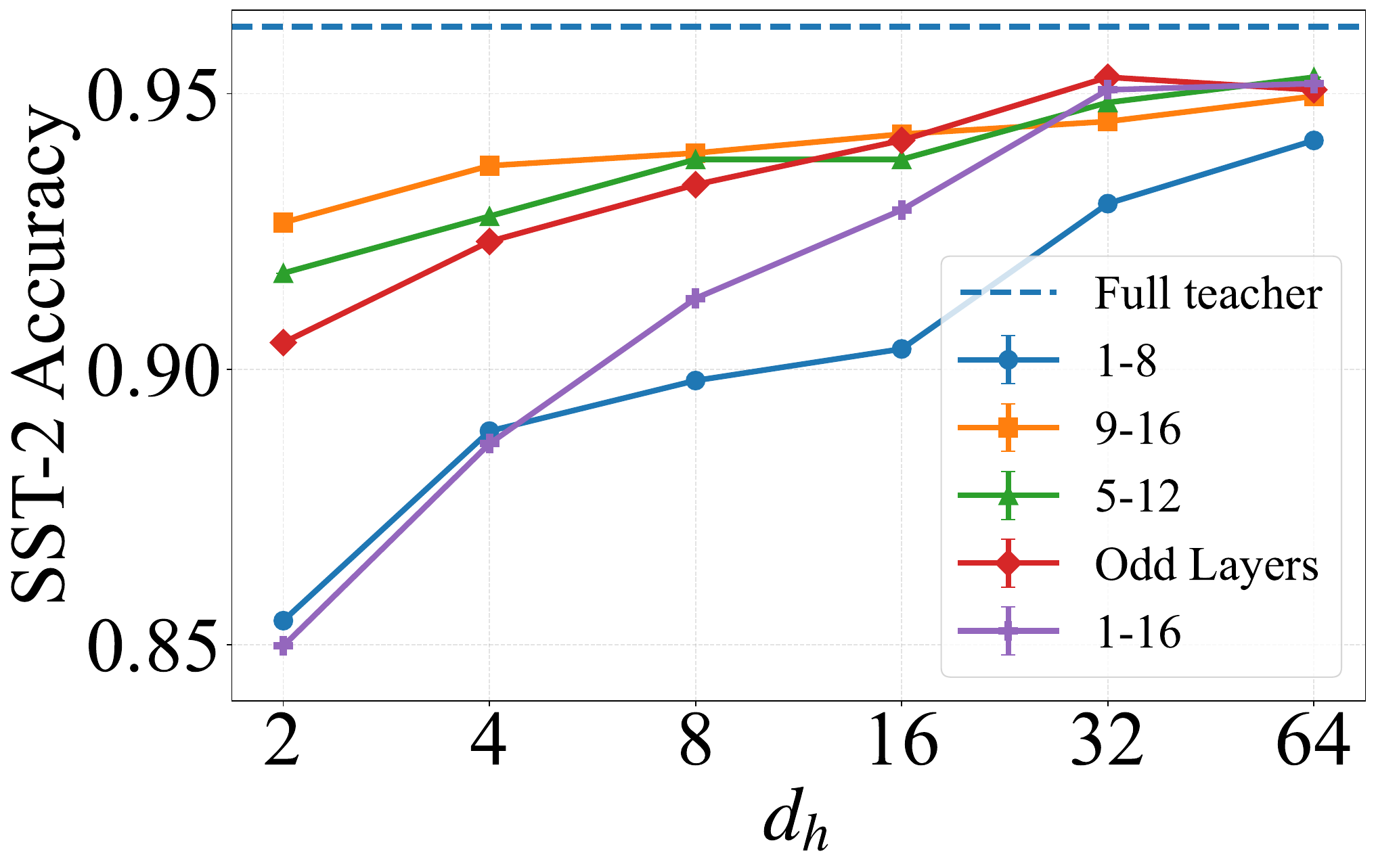}
        \caption{Reduced head dimensions in \textsc{LLaMA}-3.2-1B yield saturated SST-2 accuracy after distillation and fine-tuning.\protect\footnotemark[3]}
        \label{exp: two stage}
    \end{minipage}
    \hspace{2mm}
    \begin{minipage}[t]{0.475\linewidth}
        \centering
        \includegraphics[width=\linewidth]{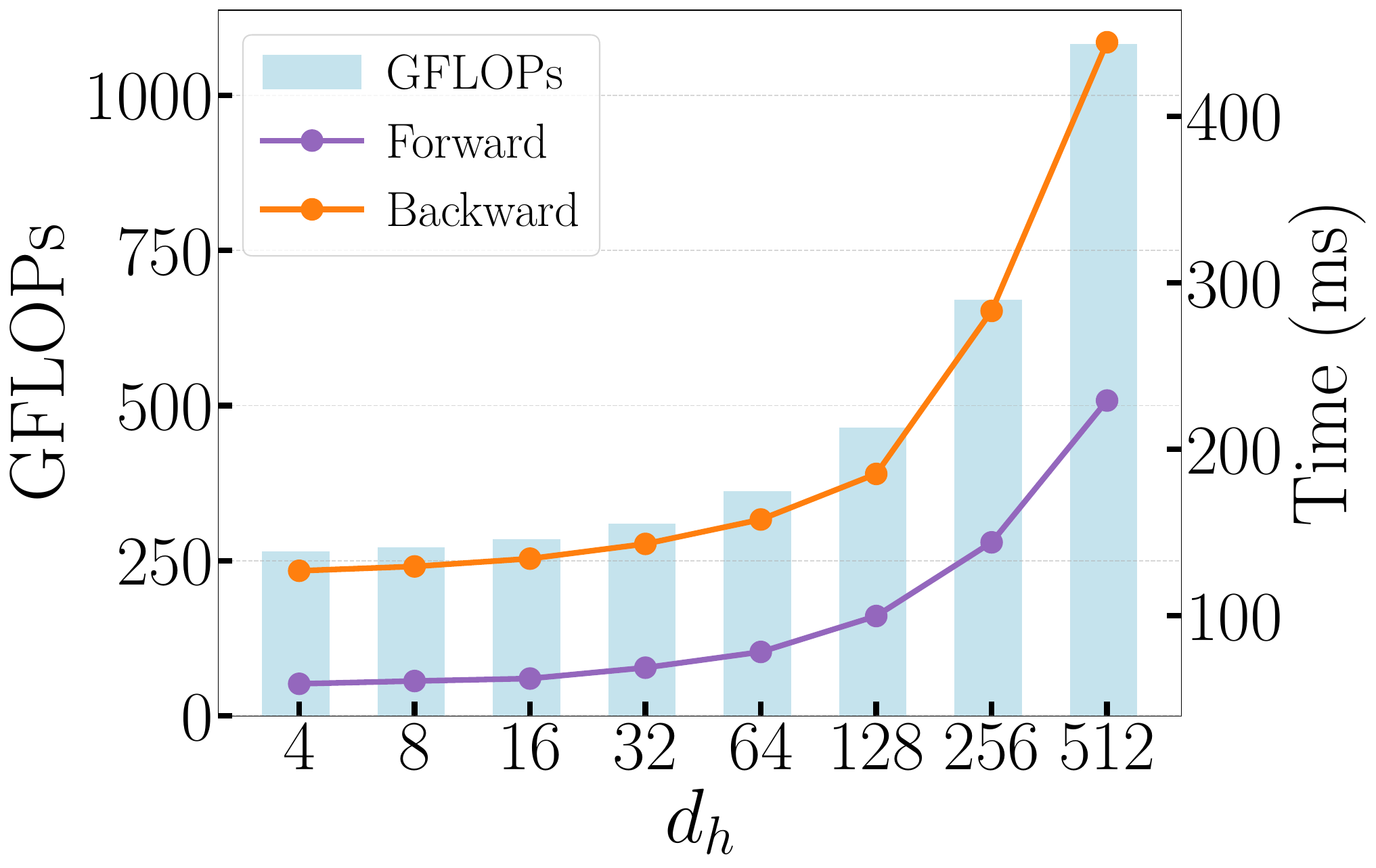}
        \caption{Computation efficiency evaluations of Transformers regarding head dimensions.} 
        \label{exp: efficiency} 
    \end{minipage}
\end{figure}

\paragraph{\textsc{LLaMA}-3.2-1B: Saturated downstream accuracy.}
We further evaluate head-dimension reduction on downstream tasks using
\textsc{LLaMA}-3.2-1B, whose original head dimension is $d_H=64$.
For selected layers (early 1--8, middle 5--12, late 9--16, odd, and
all 1--16), each query/key projection head is permanently
reparameterized as a rank-$d_h$ factorization,
\[
    W_h=L_hR_h,\qquad \operatorname{rank}(W_h)\le d_h,
\]
with $d_h\in\{2,4,8,16,32,64\}$, while all other model components are
kept unchanged. The low-rank factors are initialized using the truncated
SVD of the corresponding pretrained Q/K projections. Training consists
of two stages. In Stage I, the compressed model is refined against the
uncompressed language-model teacher on WikiText-2 for 400 steps by
minimizing a combination of hidden-state MSE and output-logit KL
divergence, while only the low-rank Q/K factors are optimized. In
Stage II, a full-rank SST-2 classifier is used as the teacher to distill
the compressed model for three epochs, with an objective combining
label cross-entropy, output KL divergence, and hidden-state matching;
only the low-rank Q/K factors and the classification head are trainable.
As shown in Figure~\ref{exp: two stage}, the compressed models retain
strong downstream performance even under substantial reductions of the
Q/K head dimension, while the degree of robustness depends noticeably
on which layers are compressed, indicating non-uniform redundancy across
depth.

\paragraph{Ablation study.} To further isolate the contribution of each component in this pipeline,
we conduct controlled ablations by fixing the compressed layers to
5--12 and $d_h=16$, with complete results reported in
Appendix~\ref{appendix: two stage ablation}. Replacing the SVD
initialization with Kaiming initialization increases the post-refinement
WikiText-2 loss from $3.215$ to $3.582$ and decreases SST-2 accuracy
from $93.81\%$ to $91.74\%$. Removing Stage I refinement decreases the
downstream accuracy from $93.81\%$ to $91.86\%$, whereas removing
Stage II downstream distillation results in only $47.48\%$ accuracy.
These comparisons demonstrate complementary roles of the three
components: truncated-SVD initialization provides a stronger
low-rank starting point, intermediate refinement better preserves the
pretrained representation after compression, and downstream
distillation is essential for recovering task-specific performance.

\paragraph{Computation efficiency analysis.}
To assess the improvement of model computation efficiency caused by head dimension reduction, we test for a customized Transformer model, which is a 6-layer encoder with the embedding dimension $D=256$ and $H=8$ attention heads. We vary the head dimension $d_h \in \{4, 8, 16, 32, 64, 128\}$ to calculate the computation overhead  scaling. 
Here, we use synthetic sequences with the sequence length $L=256$ and the batch size $B=256$ to ensure a compute-bound regime. To measure realistic training overhead, for the classification task, we run 400 optimization steps using the AdamW optimizer with a learning rate $2 \times 10^{-4}$ to minimize a cross-entropy loss.
Benchmarks are evaluated on an NVIDIA A100 GPU (80GB). The FLOPs are calculated via \texttt{thop} and averaged over 100 iterations. 
As indicated in Figure~\ref{exp: efficiency}, both GFLOPs and empirical forward/backward time scale non-linearly with $d_h$. The results confirm that low head dimensions can help to significantly reduce forward/backward time and model calculations, offering a path to more  efficient inference and training.


\footnotetext[3]{The performance gap stems from the varied tuning capacity, as the student model is partially fine-tuned only for the selected layer given reduced head dimensions, while the teacher model is fully fine-tuned. 
}




\section{Conclusion and Limitations}

In this work, we study performance–efficiency trade-offs in Transformers from an approximation theory perspective, focusing on the allocation of attention heads and head dimensions. 
Under a fixed parameter budget, we characterize an intrinsic trade-off between the number of heads and their dimensions, and additionally show that softmax-based attention exhibits saturation, where increasing head dimensions yields diminishing returns, especially for long sequences. 
Experiments across synthetic and language modeling tasks support these findings and provide guidance for the parameter-efficient architecture design. 
For limitations, our theoretical analysis mainly focuses on idealized settings, 
and thus may not capture all architectural details.  
Extending the analysis to more realistic architectures and intricate optimization dynamics remains promising as an important direction for future work. 




\appendix

\bibliography{example_paper}
\bibliographystyle{plainnat}

\clearpage
\onecolumn
\appendix
\renewcommand{\thesection}{\Alph{section}}

\begin{center}
    \noindent\rule{\textwidth}{4pt} \vspace{-0.2cm}
    \LARGE \textbf{Appendix}
    \noindent\rule{\textwidth}{1.2pt}
\end{center}

\addtocontents{toc}{\protect\setcounter{tocdepth}{2}} 
\tableofcontents

\vspace{1.cm}

\section{Use of Large Language Models}\label{appendix: clarification}

In the preparation of this manuscript, we employed a large language model (ChatGPT, developed by OpenAI) as a writing assistant. The model was used exclusively to:

\begin{itemize}
    \item Improve the clarity, coherence, and conciseness of the text.
    \item Rephrase sentences to enhance grammatical correctness and readability.
    \item Ensure consistent terminology and smooth transitions across sections.
\end{itemize}

All outputs generated by the LLM were carefully reviewed, edited, and validated by the authors to guarantee technical accuracy, logical consistency, and fidelity to the research content. No text was included without human oversight. Importantly, the LLM was not employed for data generation, model training, or experimental design.

\vspace{1.cm}

\section{Related Works}

\paragraph{Approximation ability of Transformers.} 
The work~\cite{yun2019transformers} is the first to establish the universal approximation property of Transformers. 
Subsequent studies analyzed the efficiency of Transformers in representing specific functions or tasks, including sparse functions~(\cite{edelman2022inductive}), targets with nonlinear temporal kernels~(\cite{jiang2024approximation}), practical computer programs~(\cite{giannou2023looped}), long but sparse memories~(\cite{wang2024understandingexpressivepowermechanisms}), induction heads~(\cite{sanford2024onelayertransformersfailsolve,sanford2024transformersparallelcomputationlogarithmic,rajaraman2024transformersmarkovdataconstant}), and memorization and reasoning~(\cite{chen2024transformerlearnvaryingdepth}). 
Moreover, several studies suggest that Transformers achieve in-context learning by effectively approximating gradient-based iterations across layers~(\cite{garg2022can,akyurek2022learning,von2023transformers,mahankali2023one,bai2023transformersstatisticiansprovableincontext,shen2023pretrained}). 
Other work has examined limitations of Transformers' expressivity, particularly in modeling formal languages or simulating circuits~(\cite{hahn2020theoretical,weiss2021thinking,bhattamishra2020ability,merrill2022saturated,merrill2023expresssive}). 

\paragraph{Attention structure.}
Prior work has extensively studied the representational power and structural properties of self-attention in Transformers. 
\cite{Likhosherstov_Choromanski_Weller_2023} showed that a fixed self-attention layer can approximate arbitrary sparse patterns with hidden size scaling logarithmically in sequence length.
Several works proposed replacing standard attention with linearized variants for improved efficiency (\cite{wang2020linformer,choromanski2020rethinking}).
From a rank perspective, \cite{bhojanapalli2020low} addressed the low-rank bottleneck by setting head dimension proportional to sequence length, while \cite{amsel2024quality} demonstrated that a single full-rank head can implement nearest-neighbor search, whereas exponentially many low-rank heads are required otherwise.
Finally, \cite{kajitsuka2023transformers} established memorization capacity for one-layer, single-head Transformers, whereas \cite{hahn2020theoretical} showed that modeling certain formal languages requires the number of layers or heads to grow with input length.

\paragraph{Property of softmax.}
Recent studies have questioned the suitability of softmax attention and analyzed its theoretical limitations.
\cite{mudarisov2025limitationsnormalizationattentionmechanism} showed that as sequence length increases, softmax attention collapses toward uniformity, token separability saturates, and gradients become unstable, consistent with empirical observations in GPT-2.
From a different perspective, \cite{saratchandran2024rethinking} argued that the effectiveness of softmax stems mainly from implicit Frobenius norm regularization rather than its probabilistic interpretation, and demonstrated that polynomial activations can achieve comparable performance.
Several alternatives have been proposed to address these issues, including SA-Softmax to mitigate gradient vanishing (\cite{zheng2025selfadjust}), sigmoid attention with improved sample efficiency (\cite{yan2025sigmoidattention}), and Scalable-Softmax, which preserves sharper attention distributions for long contexts (\cite{nakanishi2025scalable}).

\paragraph{Model efficiency.}
Multi-head attention (MHA) is fundamental to Transformers, yet extensive empirical evidence reveals significant redundancy across attention heads.
A large fraction of heads can be pruned with negligible performance loss, and only a small subset consistently captures meaningful patterns across tasks and languages (\cite{michel2019sixteen,voita2019analyzing,ma2021contributions}).
Building on this observation, subsequent work proposes principled and automated pruning criteria that substantially improve parameter efficiency while preserving accuracy (\cite{choi2025entropy,lee2024automatic}).
In parallel, efficiency has been further enhanced through architectural and system-level modifications, including sparse and structured attention for long contexts (\cite{beltagy2020longformer,zaheer2020bigbird}), shared key--value projections (\cite{ainslie2023gqa}), and optimized attention kernels (\cite{dao2022flashattention,hatamizadeh2023fastervit}).
Despite these advances, a theoretical understanding of when and why multiple attention heads are necessary remains largely open.

\vspace{1.cm}

\section{Lemmas used in Proofs}

The first lemma provides a convergence rate for approximating the delta function using exponential sums.

\begin{lemma}\label{lemma E.1}
For any $T\in\mathbb{N}_+$, $q,m\in\mathbb{N}_+$, there exists a $\phi_{m}^{\exp}(t)=\sum\limits_{k=1}^m \alpha_k e^{-\beta_k t}$ such that
\begin{align*}
    \|{\mathbb{I}(\cdot=T)-\phi_{m}^{\rm exp}(\cdot)}\|_{\ell_1(\mathbb{N})}
    \leq \frac{1.3\  e^{0.02T}}{m}
\end{align*}
where $\beta_k>0$ holds for any $k\in[m]$, and $A,C>0$ are absolute constants.
\end{lemma}

\begin{proof}
This lemma is a corollary of Lemma F.1 in ~\cite{wang2024understandingexpressivepowermechanisms}. By Lemma F.1 in ~\cite{wang2024understandingexpressivepowermechanisms} and its proof: for any $T\in\mathbb{N}_+$, $q,m\in\mathbb{N}_+$, there exists a $C(q)>0$ and a $\phi_{m}^{\exp}(t)=\sum\limits_{k=1}^m \alpha_k e^{-\beta_k t}$ such that
\begin{align*}
    \|{\mathbb{I}(\cdot=T)-\phi_{m}^{\rm exp}(\cdot)}\|_{\ell_1(\mathbb{N})}
    \leq\frac{C(q)e^{0.01(q+1)T}}{m^{q}},
\end{align*}
where $\beta_k>0$ holds for any $k\in[m]$. Moreover, 
\begin{align*}
    C(q)=\frac{M(q)}{(1-1/e)^q},\quad 
    M(q)=\max_{0\leq k\leq q}\sup_{x\in[-1,1]}\left|\Psi^{(k)}(x)\right|,
\end{align*}
where $\Psi(x)=\begin{cases}
    \exp\left(-\frac{1}{1-x^2}\right),\ x\in(-1,1) \\
    0,\ \mathrm{otherwise}
\end{cases}$ is the standard bump function on $[-1,1]$. We mainly focus on small $q$ since $C(q)$ grows rapidly and through directly computation, we have $C(1)\approx 1.3$ . Hence,
\begin{align*}
    \|{\mathbb{I}(\cdot=T)-\phi_{m}^{\rm exp}(\cdot)}\|_{\ell_1(\mathbb{N})}
    \leq \frac{1.3\  e^{0.02T}}{m}
\end{align*}
\end{proof}

The following lemma is for technical usage.
\begin{lemma}[Tensor-product perturbation]
\label{lemma: tensor error}
Let $u_1,\ldots,u_n$ and $v_1,\ldots,v_n$ be vectors satisfying
\[
    \|u_j\|_2\le B,
    \qquad
    \|u_j-v_j\|_2\le \varepsilon,
    \qquad j\in[n].
\]
Then
\[
\left\|
    u_1\otimes\cdots\otimes u_n
    -
    v_1\otimes\cdots\otimes v_n
\right\|_2
\le
(B+\varepsilon)^n-B^n.
\]
\end{lemma}

\begin{proof}
Since $\|v_j\|_2\le B+\varepsilon$, the telescoping identity gives
\[
\begin{aligned}
&u_1\otimes\cdots\otimes u_n
-v_1\otimes\cdots\otimes v_n \\
&=
\sum_{j=1}^n
v_1\otimes\cdots\otimes v_{j-1}
\otimes (u_j-v_j)
\otimes u_{j+1}\otimes\cdots\otimes u_n.
\end{aligned}
\]
Therefore,
\[
\begin{aligned}
\left\|
    u_1\otimes\cdots\otimes u_n
    -v_1\otimes\cdots\otimes v_n
\right\|_2
&\le
\sum_{j=1}^n
(B+\varepsilon)^{j-1}\varepsilon B^{n-j} \\
&=(B+\varepsilon)^n-B^n.
\end{aligned}
\]
\end{proof}

The following lemma guarantees that, with high probability, a bounded random vector can be projected onto a $d$-dimensional subspace such that the resulting error, after applying any linear transformation $A$, is tightly controlled by $\|A\|_2$, the projection dimension, and the vector’s norm bound.

\begin{lemma}[Uniform high-probability projection for bounded random vectors]
\label{lemma: prob projection}
Let
\[
    X_1,\ldots,X_N\in\mathbb{R}^{D}
\]
be random vectors defined on a common probability space. They are not
required to be independent, identically distributed, isotropic, or
sub-Gaussian. Assume only that
\[
    \|X_\ell\|_2\le B
    \qquad\text{almost surely for all }\ell\in[N].
\]

Let
\[
    E\sim \mathrm{Unif}(G_{D,r})
\]
be a uniformly random $r$-dimensional subspace of $\mathbb{R}^{D}$,
independent of $(X_1,\ldots,X_N)$, and let $\Pi_E$ denote the
orthogonal projection onto $E$. Equivalently, let
$P\in\mathbb{R}^{r\times D}$ have orthonormal rows spanning $E$, so that
\[
    PP^\top=I_r,
    \qquad
    P^\top P=\Pi_E.
\]

Then there exists an absolute constant $c>0$ such that, for every
$\delta\in(0,1)$, with probability at least $1-\delta$ over the joint
randomness of $(X_1,\ldots,X_N,E)$,
simultaneously for all $\ell\in[N]$,
\[
    \|(I-\Pi_E)X_\ell\|_2
    \le
    B\left(
        \sqrt{1-\frac rD}
        +
        \sqrt{
            \frac{\log(2N/\delta)}
                 {cD}
        }
    \right).
\]

More strongly, conditional on $(X_1,\ldots,X_N)$, the above event has
probability at least $1-\delta$ with respect to the randomness of $E$
almost surely.

Moreover, for every $\ell\in[N]$,
\[
    \mathbb{E}_{E}
    \left[
        \left.
        \|(I-\Pi_E)X_\ell\|_2^2
        \right|X_\ell
    \right]
    =
    \left(1-\frac rD\right)
    \|X_\ell\|_2^2
    \qquad\text{almost surely},
\]
and consequently
\[
    \mathbb{E}_{X_\ell,E}
    \|(I-\Pi_E)X_\ell\|_2^2
    =
    \left(1-\frac rD\right)
    \mathbb{E}\|X_\ell\|_2^2.
\]

When $r=D$, the projection error is identically zero.
\end{lemma}

\begin{proof}
Let
\[
    Q_E:=I-\Pi_E,
\]
which is the orthogonal projection onto the $(D-r)$-dimensional
subspace $E^\perp$.

We first establish the concentration inequality for a fixed
deterministic vector $x\in\mathbb{R}^D$. If $x=0$, the result is
trivial. Hence suppose $x\neq 0$ and write
\[
    u=\frac{x}{\|x\|_2}\in\mathbb{S}^{D-1}.
\]

By rotational invariance of the uniform measure on the Grassmannian,
the random variable
\[
    \|Q_Eu\|_2
\]
has the same distribution as
\[
    \|Q_0U\|_2,
\]
where $U$ is uniformly distributed on the unit sphere
$\mathbb{S}^{D-1}$ and $Q_0$ is any fixed orthogonal projection of rank
$D-r$.

Consider the function
\[
    f(u):=\|Q_0u\|_2,
    \qquad
    u\in\mathbb{S}^{D-1}.
\]
Since $Q_0$ is an orthogonal projection,
\[
    |f(u)-f(v)|
    \le
    \|Q_0(u-v)\|_2
    \le
    \|u-v\|_2,
\]
so $f$ is $1$-Lipschitz.

By the standard concentration inequality for Lipschitz functions on
the sphere, there exists an absolute constant $c>0$ such that, for
every $\varepsilon>0$,
\[
    \Pr\left(
        f(U)\ge \mathbb{E}f(U)+\varepsilon
    \right)
    \le
    2\exp(-cD\varepsilon^2).
\]
On the other hand, Jensen's inequality gives
\[
    \mathbb{E}f(U)
    \le
    \sqrt{\mathbb{E}f(U)^2}.
\]
Using
\[
    \mathbb{E}[UU^\top]=\frac1D I_D,
\]
we obtain
\[
    \mathbb{E}f(U)^2
    =
    \mathbb{E}\|Q_0U\|_2^2
    \leq
    \operatorname{Tr}
    \left(
        Q_0\mathbb{E}[UU^\top]
    \right)
    =
    \frac{\operatorname{Tr}(Q_0)}{D}
    =
    1-\frac rD.
\]
Therefore,
\[
    \mathbb{E}f(U)
    \le
    \sqrt{1-\frac rD}.
\]
It follows that
\[
    \Pr_E\left[
        \|Q_Eu\|_2
        \ge
        \sqrt{1-\frac rD}+\varepsilon
    \right]
    \le
    2\exp(-cD\varepsilon^2).
\]
Multiplying by $\|x\|_2$ yields
\begin{equation}\label{equation: lemma proj}
        \Pr_E\left[
        \|Q_Ex\|_2
        \ge
        \|x\|_2
        \left(
            \sqrt{1-\frac rD}+\varepsilon
        \right)
    \right]
    \le
    2\exp(-cD\varepsilon^2).
    \tag{1}
\end{equation}

We now return to the random vectors
$X_1,\ldots,X_N$. Let
\[
    \mathcal{F}_X
    :=
    \sigma(X_1,\ldots,X_N).
\]
Since $E$ is independent of $\mathcal{F}_X$, conditional on
$\mathcal{F}_X$ the subspace $E$ is still uniformly distributed over
$G_{D,r}$.

Fix an arbitrary realization
\[
    X_1=x_1,\ldots,X_N=x_N
\]
satisfying
\[
    \|x_\ell\|_2\le B,
    \qquad \ell\in[N].
\]
Applying~\eqref{equation: lemma proj} to each $x_\ell$ gives
\[
    \Pr_E\left[
        \|Q_Ex_\ell\|_2
        \ge
        B
        \left(
            \sqrt{1-\frac rD}+\varepsilon
        \right)
    \right]
    \le
    2\exp(-cD\varepsilon^2),
\]
where we used $\|x_\ell\|_2\le B$.

Hence, by a union bound over $\ell\in[N]$,
\[
\begin{aligned}
    &\Pr_E\left[
        \max_{\ell\in[N]}
        \|Q_Ex_\ell\|_2
        \ge
        B
        \left(
            \sqrt{1-\frac rD}+\varepsilon
        \right)
        \,\middle|\,
        X_1=x_1,\ldots,X_N=x_N
    \right]
    \\
    &\hspace{3cm}
    \le
    2N\exp(-cD\varepsilon^2).
\end{aligned}
\]
Since this bound holds for every realization satisfying the almost
sure boundedness assumption, we have almost surely
\[
    \Pr_E\left[
        \left.
        \max_{\ell\in[N]}
        \|Q_EX_\ell\|_2
        \ge
        B
        \left(
            \sqrt{1-\frac rD}+\varepsilon
        \right)
        \right|
        \mathcal{F}_X
    \right]
    \le
    2N\exp(-cD\varepsilon^2).
\]
Taking expectation with respect to $\mathcal{F}_X$ gives
\[
    \Pr_{X,E}\left[
        \max_{\ell\in[N]}
        \|Q_EX_\ell\|_2
        \ge
        B
        \left(
            \sqrt{1-\frac rD}+\varepsilon
        \right)
    \right]
    \le
    2N\exp(-cD\varepsilon^2).
\]
Choosing
\[
    \varepsilon
    =
    \sqrt{
        \frac{\log(2N/\delta)}
             {cD}
    }
\]
proves the uniform high-probability bound.

For the expectation identity, rotational invariance of the uniform
distribution on the Grassmannian implies
\[
    \mathbb{E}_E[\Pi_E]
    =
    \frac rD I_D.
\]
Consequently,
\[
    \mathbb{E}_E[Q_E]
    =
    \left(1-\frac rD\right)I_D.
\]
Conditioning on $X_\ell$, we obtain
\[
\begin{aligned}
    \mathbb{E}_E
    \left[
        \left.
        \|Q_EX_\ell\|_2^2
        \right|X_\ell
    \right]
    &=
    X_\ell^\top
    \mathbb{E}_E[Q_E^\top Q_E]
    X_\ell
    \\
    &=
    X_\ell^\top
    \mathbb{E}_E[Q_E]
    X_\ell
    \\
    &=
    \left(1-\frac rD\right)
    \|X_\ell\|_2^2,
\end{aligned}
\]
where we used $Q_E^2=Q_E$. Taking expectation over $X_\ell$ yields
\[
    \mathbb{E}_{X_\ell,E}
    \|Q_EX_\ell\|_2^2
    =
    \left(1-\frac rD\right)
    \mathbb{E}\|X_\ell\|_2^2.
\]

Finally, when $r=D$, we have $E=\mathbb{R}^D$ and hence
$Q_E=0$, so the projection error is identically zero.
\end{proof}

The following lemma is Corollary 1 in ~\cite{mudarisov2025limitationsnormalizationattentionmechanism}.

\begin{lemma}\label{lemma: re scale}
    Assume $\|\mathrm{Logits}_{d_H}(\boldsymbol{W}_Q,\boldsymbol{W}_K,\boldsymbol{X})\|_{\infty}$ is bounded, the attention weights satisfy:
    \begin{equation*}
        \frac{1}{L}\exp\left(-2\Delta\right)\leq \|\mathrm{AttnScore}_{d_H}(\boldsymbol{W}_Q,\boldsymbol{W}_K,\boldsymbol{X})\|_{\infty}\leq \frac{1}{L}\exp\left(2\Delta\right),
    \end{equation*}
    where $\Delta=\|\boldsymbol{W}_Q\|_2\|\boldsymbol{W}_K\|_2B^2$.
\end{lemma}

The Gershgorin circle theorem localizes all eigenvalues of a matrix within discs determined by its diagonal entries and row sums of off-diagonal magnitudes.

\begin{lemma}[Gershgorin circle theorem]\label{lemma: gershgorin}
Let $A=(a_{ij})\in\mathbb{C}^{n\times n}$. For each $i$ define the Gershgorin radius
\[
R_i:=\sum_{j\neq i}|a_{ij}|.
\]
Then every eigenvalue $\lambda$ of $A$ lies in the union of the Gershgorin discs
\[
\lambda\in\bigcup_{i=1}^n\{z\in\mathbb{C}:\ |z-a_{ii}|\le R_i\}.
\]
In particular, if the discs $\{z:|z-a_{ii}|\le R_i\}$ are contained in a region $S\subset\mathbb{C}$, then all eigenvalues of $A$ lie in $S$.
\end{lemma}

The mean value theorem expresses the difference 
$F(\boldsymbol{y}) - F(\boldsymbol{x})$ as the Jacobian at an intermediate point applied to 
$\boldsymbol{y}-\boldsymbol{x}$.

\begin{lemma}[Mean Value Inequality for vector-valued functions]\label{lemma: vector mean value}
Let $F:\mathbb{R}^n \to \mathbb{R}^m$ be continuously differentiable on an open convex set $D \subset \mathbb{R}^n$. 
Then for any $\boldsymbol{x}, \boldsymbol{y} \in D$, there exists $\theta \in (0,1)$ such that
\begin{equation*}
    \| F(\boldsymbol{y}) - F(\boldsymbol{x}) \| \le \| JF(\boldsymbol{x} + \theta (\boldsymbol{y} - \boldsymbol{x})) \| \cdot \| \boldsymbol{y} - \boldsymbol{x} \|,
\end{equation*}
where $JF(\cdot)$ denotes the Jacobian matrix of $F$ and $\|\cdot\|$ is an appropriate vector/matrix norm.
\end{lemma}

\begin{lemma}[Eckart--Young--Mirsky Theorem]\label{lemma: eckart young}
Let $\boldsymbol{A} \in \mathbb{R}^{m \times n}$ have singular value decomposition
$\boldsymbol{A} = \sum_{i=1}^{r} \sigma_i \boldsymbol{u}_i \boldsymbol{v}_i^\top$,
where $\sigma_1 \ge \sigma_2 \ge \cdots \ge \sigma_r > 0$.
For any $k < r$, the rank-$k$ truncated SVD
\[
\boldsymbol{A}_k := \sum_{i=1}^{k} \sigma_i \boldsymbol{u}_i \boldsymbol{v}_i^\top
\]
is an optimal rank-$k$ approximation of $\boldsymbol{A}$ in both the spectral norm and the Frobenius norm, i.e.,
\[
\boldsymbol{A}_k \in \arg\min_{\operatorname{rank}(\boldsymbol{B}) \le k}
\|\boldsymbol{A} - \boldsymbol{B}\|_2
\quad \text{and} \quad
\boldsymbol{A}_k \in \arg\min_{\operatorname{rank}(\boldsymbol{B}) \le k}
\|\boldsymbol{A} - \boldsymbol{B}\|_F.
\]
Moreover, the optimal errors satisfy
\[
\|\boldsymbol{A} - \boldsymbol{A}_k\|_2 = \sigma_{k+1},
\qquad
\|\boldsymbol{A} - \boldsymbol{A}_k\|_F^2 = \sum_{i=k+1}^{r} \sigma_i^2 .
\]
\end{lemma}

\vspace{1.cm}

\section{Proof of Theorem \ref{theorem: tradeoff}}\label{appendix: proof1}

In this section, we give the detailed proof of Theorem \ref{theorem: tradeoff}.  We first restate the theorem:

\begin{theorem}[Restate of Theorem \ref{section: tradeoff}.]\label{theorem: re tradeoff}
    To approximate the linear target $\mathrm{H}_t(\boldsymbol{X})=\sum_{i=0}^{L}\boldsymbol{\rho}_i\boldsymbol{x}_{t-i}$,
    suppose we have $H_m$ heads each of the dimension $d_m$ (m=1,\ldots,M), and the model dimension $D=\sum_{m=1}^M H_m\cdot d_m$ is fixed. Then, with probability at least $1-\delta$, we have
    \begin{equation}\label{equation: re E(D)}
    \begin{aligned}
    \mathcal{E}_D(X)\le& \quad B\sum_{m=1}^{M}\|\rho_m\|_2
    \left[\mathbb{I}_{\{d_m<d\}}\left(\sqrt{1-\frac{d_m}{d}}\right)+\frac{1.3e^{0.02m}}{H_m}\right]+B\sum_{k=M+1}^{L}\|\rho_k\|_2\\
    &+B\sum_{m=1}^{M}\|\rho_m\|_2 \cdot C\sqrt{\frac{\log(2M/\delta)}{d}}.
    \end{aligned}
    \end{equation}
    \end{theorem}

\begin{proof}

We only consider the case $d_m \leq d$ in our proof, since otherwise there is no information loss and the first term in the approximation error vanishes. Since we only use a single layer of the Transformer, we omit the layer index.

\paragraph{Embedding layer.} 

We embed each token by
$$\boldsymbol{W}_E = \begin{pmatrix}
    \boldsymbol{I}_d\\
    \boldsymbol{0}_{(D-d)\times d}
\end{pmatrix}\in\mathbb{R}^{D\times d},\quad \boldsymbol{b}_E=\boldsymbol{0}\in\mathbb{R}^D,$$
for $\boldsymbol{x}_t$, then each token $\boldsymbol{x}_t^{(0)}$ after embedding is
$$\boldsymbol{x}_t^{(0)} = \boldsymbol{W}_E\boldsymbol{x}_t + \boldsymbol{b}_E = \begin{pmatrix}
    {\boldsymbol{x}}_t\\\boldsymbol{0}
\end{pmatrix}\in\mathbb{R}^{D}.$$

\paragraph{Multi-head attention layer with residual connection.} We implement multi-head attention using $M$ groups of heads. Each group $m$ comprises $H_m$ heads with dimension $d_m$ for $(m=1,2,\ldots, M)$, aimed at extracting the corresponding representation $\tilde{\boldsymbol{x}}_{t-m}$ which is ${\boldsymbol{x}}_{t-m}$ after  projection to $d_m$-dimension space.
This step can be presented as follows for each token:
\begin{equation*}
    \begin{pmatrix}
        {\boldsymbol{x}}_t\\
        0\\
        \vdots\\
        0
    \end{pmatrix}\longrightarrow
    \begin{pmatrix}
        \tilde{\boldsymbol{x}}_t\\
        \tilde{\boldsymbol{x}}_{t-1}\\
        \vdots\\
        \tilde{\boldsymbol{x}}_{t-M}\\
        \boldsymbol{0}_{D-(\sum_{m=1}^M d_m)}
    \end{pmatrix}:=\boldsymbol{x}_t^{(1/2)}.
\end{equation*}
where $\tilde{\boldsymbol{x}}_t$ is obtained by residual connection.

We now focus on the extraction of an arbitrary $\tilde{\boldsymbol{x}}_{t-m}$ and give the details of construction.

By lemma \ref{lemma E.1}, for any rate $q\in\mathbb{N}^+$, there exists a function 
$$\phi_m^{\exp}(t)=\sum_{h=1}^{H_m}\alpha_{h,m} e^{-\beta_{h,m} (t-1)}$$
such that $\beta_h>0$ and 
$$\Vert \mathbb{I}\left(\cdot=m\right)-\phi_m^{\exp}(\cdot) \Vert_{\ell_1(\mathbb{N})}=\sum_{s=0}^{+\infty}|\mathbb{I}\left(s=m\right)-\phi^{\exp}(s)|\leq \frac{1.3\  e^{0.02m}}{H_m}.
$$

By Lemma \ref{lemma: prob projection}, there exist an projection $\boldsymbol{P}_m\in\mathbb{R}^{d_m\times d}$ such that
$$
\|(I-\Pi_E)x_\ell\|_2
    \le
    B\sqrt{1-\frac {d_m}D}
        +B\cdot\sqrt{
        \frac{\log(2N/\delta)}
        {cD}
        }
$$
for any $l=1,\ldots,m$ with probability at least $1-\delta$. Let $\tilde{\boldsymbol{x}}_{t-m}:=\boldsymbol{P}_m\boldsymbol{x}_{t-m}$
For $h=\sum_{i=1}^{m-1}H_i,\ \sum_{j=1}^{m-1}H_j+1\ ,\ldots,\ \sum_{k=1}^{m}H_k$, we choose parameters as follows
$$p^{(h)}=\beta_{h,m},\quad \boldsymbol{W}_V^{(h)}=\alpha_{h,m}\left(\sum_{i=0}^{H_m}\exp(-\beta_{h,m} (i-1))\right)\boldsymbol{P}_m,\quad\boldsymbol{W}_{K}^{(h)}=\boldsymbol{W}_{Q}^{(h)}={0}.$$

The output of $H$ heads are concatenated together, and we take 
$$\boldsymbol{W}_O=\sum_{m=1}^{M}\sum_{h=\sum_{i=1}^{m-1}H_i+1}^{\sum_{j=1}^{m}H_j}\boldsymbol{S}_{m,h},$$
where $\boldsymbol{S}_{m,h}\in\mathbb{R}^{D\times D}$ moves the output of the $m-$th set of heads to the same position
$$\boldsymbol{S}_{m,h}=\begin{pmatrix}
    \boldsymbol{0}_{(\sum_{i=1}^{m-1}d_i)\times(\sum_{i=1}^{m-1}d_i+(h-1)d_m)} &\boldsymbol{0}_{(\sum_{i=1}^{m-1}d_i)\times d_m} &\boldsymbol{0}_{(\sum_{i=1}^{m-1}d_i)\times(D-\sum_{i=1}^{m-1}d_i-hd_m)}\\
    \\
    \boldsymbol{0}_{d_m\times(\sum_{i=1}^{m-1}d_i+(h-1)d_m)}&\boldsymbol{I}_{d_m\times d_m} &\boldsymbol{0}_{d_m\times(D-\sum_{i=1}^{m-1}d_i-hd_m)}\\
    \\
    \boldsymbol{0}_{(D-\sum_{i=1}^m d_m)\times(\sum_{i=1}^{m-1}d_i+(h-1)d_m)}& \boldsymbol{0}_{(D-\sum_{i=1}^m d_m)\times d_m}& \boldsymbol{0}_{(D-\sum_{i=1}^m d_m)\times(D-\sum_{i=1}^{m-1}d_i-hd_m)}
\end{pmatrix} $$

We define
$$\tilde{\boldsymbol{P}}_m:=\left(0_{d_H\times(m-1)d_H},I_{d_m\times d_m},0_{d_H\times(D-md_H)}\right)\in\mathbb{R}^{d_m\times D}, \quad m\in[M],$$
then
\begin{align*}
	\tilde{\boldsymbol{P}}_{m}\cdot\mathrm{Concat}\bigg(\mathrm{Attn}^{(h)}(\boldsymbol{X}^{(0)})\bigg)_{h=1}^H=\sum_{h=\sum_{i=1}^{m-1}H_i}^{\sum_{j=1}^{m}H_j}\alpha_{h,m}\sum_{s=1}^{\infty}e^{-\beta_{h,m} (s-1)}\tilde{\boldsymbol{x}}_{t-s}:=\hat{\boldsymbol{x}}_{t-m}.
\end{align*}

And the approximation error of this step is 
\begin{align*}
	\varepsilon_{\mathrm{Attn}}
    \leq&\sup_t\sum_{m=1}^{M}\left\|\tilde{\boldsymbol{x}}_{t-m}-\hat{\boldsymbol{x}}_{t-m}\right\|_{\infty}
    \\\leq&\sup_t\sum_{m=1}^{M}\Vert \mathbb{I}\left(\cdot=m\right)-\phi_m^{\exp}(\cdot) \Vert_{\ell_1(\mathbb{N})}\cdot B
    \\\leq& B\cdot
        \sum_{m=1}^{M}\frac{1.3\  e^{0.02m}}{H_m}, 
\end{align*}

\paragraph{Feed forward network.} FFN is implemented component-wise to realize convolution. The final output for the $t$-th token is
$$\hat{H}_t(\boldsymbol{X})=\sum_{m=0}^{M}\boldsymbol{\tilde{\rho}}_m\hat{\boldsymbol{x}}_{t-m},$$
where $\boldsymbol{\tilde{\rho}}_m=\boldsymbol{\rho}_m{\boldsymbol{P}_m}^\top$. In fact we only need a linear map to achieve this, and FFN won't cause any approximation error.

\paragraph{Approximation error.} By Lemma \ref{lemma: prob projection}, with probability at least  
$1-\delta$, we have

\begin{equation*}
    \begin{aligned}
        \mathcal{E}_D(\boldsymbol{X})=&\|\sum_{m=0}^{\infty}\boldsymbol{{\rho}}_m{\boldsymbol{x}}_{t-m}-\sum_{m=0}^{M}\boldsymbol{\tilde{\rho}}_m\hat{\boldsymbol{x}}_{t-m}\|_2
        \\\leq&\sum_{m=1}^M\|\boldsymbol{\rho}_m\boldsymbol{x}_{t-m}-\boldsymbol{\tilde{\rho}}_m\hat{\boldsymbol{x}}_{t-m}\|_2+B\sum_{k=M+1}^{\infty}\|\boldsymbol{\rho}_k\|_2
        \\\leq&\sum_{m=1}^M\|\boldsymbol{\rho}_m\boldsymbol{x}_{t-m}-\boldsymbol{\tilde{\rho}}_m\tilde{\boldsymbol{x}}_{t-m}\|_2+\sum_{m=1}^M\|\tilde{\boldsymbol{\rho}}_m\tilde{\boldsymbol{x}}_{t-m}-\boldsymbol{\tilde{\rho}}_m\hat{\boldsymbol{x}}_{t-m}\|_2+B\sum_{k=M+1}^{\infty}\|\boldsymbol{\rho}_k\|_2
        \\\text{(By Lemma \ref{lemma: prob projection} )}\le& \quad B\sum_{m=1}^{M}\|\rho_m\|_2
        \left[\mathbb{I}_{\{d_m<d\}}\left(\sqrt{1-\frac{d_m}{d}}\right)+\frac{1.3e^{0.02m}}{H_m}\right]+B\sum_{k=M+1}^{L}\|\rho_k\|_2\\
        &+B\sum_{m=1}^{M}\|\rho_m\|_2 \cdot C\sqrt{\frac{\log(2M/\delta)}{d}}.
    \end{aligned}
\end{equation*}
\end{proof}



\vspace{1.cm}

\section{Proofs for Section~\ref{section: budget}}

\subsection{Proof of Theorem \ref{theorem: softmax} and Corollary \ref{corollary: softmax}}\label{appendix: proof2}

In this section, we prove Theorem \ref{theorem: softmax} and  Corollary \ref{corollary: softmax}, we first restate the results: 

\begin{theorem}[Restate of Theorem \ref{theorem: softmax}]
Assume
$
    \|W_Q\|_2\le M_Q,
    \,
    \|W_K\|_2\le M_K,
    \,
    \|x_t\|_2\le B\, (t\in[L]),
$
and that the finite positional-bias entries satisfy
$
    |R_{lj}|\le R_0\,(j\in\mathcal S_l,\ l\in[L]).
$
Let
$
    \Delta:=M_QM_KB^2+R_0.
$
Then, for every $l\in[L]$,
$$
    \|J_l\|_2
    \le
    \frac{2e^{2\Delta}}{s_l}.
$$
Consequently,
$$
    \|J_l\|_2=\mathcal O(L^{-1})
$$
for unmasked attention, while under the standard causal mask,
$
    \|J_l\|_2=\mathcal O(l^{-1}).
$
The hidden constants depend only on $M_Q,M_K,B$, and $R_0$.
\end{theorem}

\begin{corollary}[Restate of Theorem \ref{corollary: softmax}]
Under the assumptions of Theorem~\ref{theorem: softmax}, let
$
    M:=W_Q^\top W_K,
$
and let $M_{d_h}$ be its rank-$d_h$ truncated SVD. Choose
$\hat{W}_Q,\hat{W}_K\in\mathbb R^{d_h\times D}$ such that
$
    \hat{W}_Q^\top\hat{W}_K=M_{d_h}.
$
Let $\hat A$ denote the attention-score matrix obtained by replacing
$W_Q,W_K$ with $\hat{W}_Q,\hat{W}_K$ while keeping the same
positional bias and mask. Then for every query $l\in[L]$,
\[
    \left\|
    e_l^\top\bigl(
        \mathrm{AttnScore}_{d_H}-\hat A
    \bigr)
    \right\|_2
    \le
    \frac{2e^{2\Delta}}{s_l}
    \|X\|_2^2
    \sigma_{d_h+1}(M).
\]
Hence the per-query approximation error is
\[
    \mathcal O\!\left(
        \frac{\|X\|_2^2\sigma_{d_h+1}(M)}{L}
    \right)
\]
in the unmasked case and
\[
    \mathcal O\!\left(
        \frac{\|X\|_2^2\sigma_{d_h+1}(M)}{l}
    \right)
\]
under the standard causal mask.

In particular, for any fixed $\alpha\in(0,1)$, the late-query block
$\mathcal I_\alpha:=\{l:\,l\ge\lceil\alpha L\rceil\}$ satisfies
\[
    \left(
    \sum_{l\in\mathcal I_\alpha}
    \left\|
    e_l^\top\bigl(
        \mathrm{AttnScore}_{d_H}-\widehat A
    \bigr)
    \right\|_2^2
    \right)^{1/2}
    \le
    \frac{2e^{2\Delta}}{\alpha\sqrt L}
    \|X\|_2^2
    \sigma_{d_h+1}(M).
\]
\end{corollary}

\begin{proof}[Proof of Theorem~\ref{theorem: softmax}]
Fix $l\in[L]$ and write $p=a_l\in\mathbb R^{s_l}$. For every active
coordinate $j\in\mathcal S_l$,
\[
    |z_{lj}|
    \le
    \|W_Q\|_2\|W_K\|_2\|x_l\|_2\|x_j\|_2+|R_{lj}|
    \le \Delta.
\]
Therefore
\[
    p_j
    =\frac{e^{z_{lj}}}{\sum_{k\in\mathcal S_l}e^{z_{lk}}}
    \le
    \frac{e^\Delta}{s_le^{-\Delta}}
    =\frac{e^{2\Delta}}{s_l}.
\]
The softmax Jacobian is
\[
    J_l=\operatorname{diag}(p)-pp^\top,
\]
which is positive semidefinite. By the Gershgorin circle theorem (Lemma \ref{lemma: gershgorin}), every
eigenvalue $\lambda$ of $J_l$ satisfies
\[
    0\le\lambda
    \le
    2\max_j p_j(1-p_j)
    \le
    2\max_j p_j
    \le
    \frac{2e^{2\Delta}}{s_l}.
\]
Since $J_l$ is symmetric positive semidefinite,
\[
    \|J_l\|_2=\lambda_{\max}(J_l)
    \le\frac{2e^{2\Delta}}{s_l}.
\]
Substituting $s_l=L$ for unmasked attention and $s_l=l$ for the
standard causal mask proves the result.
\end{proof}

\begin{proof}[Proof of Corollary~\ref{corollary: softmax}]
Let $M_{d_h}$ be the rank-$d_h$ truncated SVD of $M=W_Q^\top W_K$.
By the Eckart--Young--Mirsky theorem (Lemma \ref{lemma: eckart young}),
\[
    \|M-M_{d_h}\|_2=\sigma_{d_h+1}(M),
    \qquad
    \|M_{d_h}\|_2\le\|M\|_2.
\]
Writing $M_{d_h}=U\Sigma V^\top$, one may take
\[
    \widehat W_Q=\Sigma^{1/2}U^\top,
    \qquad
    \widehat W_K=\Sigma^{1/2}V^\top,
\]
so that $\hat{W}_Q^\top\hat{W}_K=M_{d_h}$.

For query $l$, let $z_l$ and $\hat z_l$ denote the original and
compressed active logit vectors. Their difference obeys
\[
\begin{aligned}
    \|z_l-\hat z_l\|_2
    &\le
    \|X^\top(M-M_{d_h})X\|_2 \\
    &\le
    \|X\|_2^2\sigma_{d_h+1}(M).
\end{aligned}
\]
Moreover, both endpoints and every point on the line segment between
$z_l$ and $\hat z_l$ have active coordinates bounded in absolute
value by $\Delta$, because $\|M_{d_h}\|_2\le\|M\|_2$. Hence
Theorem~\ref{theorem: softmax} applies uniformly along this segment.
Using the integral form of the mean-value inequality,
\[
\begin{aligned}
    \|\operatorname{softmax}(z_l)-\operatorname{softmax}(\widehat z_l)\|_2
    &\le
    \sup_{t\in[0,1]}
    \|J_l((1-t)z_l+t\hat z_l)\|_2
    \|z_l-\hat z_l\|_2 \\
    &\le
    \frac{2e^{2\Delta}}{s_l}
    \|X\|_2^2\sigma_{d_h+1}(M).
\end{aligned}
\]
This proves the per-query bound. For
$\mathcal I_\alpha=\{l:l\ge\lceil\alpha L\rceil\}$ under causal
masking, $s_l=l\ge\alpha L$, and therefore
\[
\begin{aligned}
&\left(
    \sum_{l\in\mathcal I_\alpha}
    \|e_l^\top(\mathrm{AttnScore}_{d_H}-\hat A)\|_2^2
\right)^{1/2} \\
&\qquad\le
\sqrt L\,
\frac{2e^{2\Delta}}{\alpha L}
\|X\|_2^2\sigma_{d_h+1}(M),
\end{aligned}
\]
which gives the stated $L^{-1/2}$ late-query bound.
\end{proof}

\subsection{Extension to Rotary Positional Embeddings}
\label{appendix: rope}

We next show that the head-dimension reduction argument extends to
rotary positional embeddings (RoPE). Unlike the additive positional
bias considered in Theorem~\ref{theorem: softmax}, RoPE rotates the
query and key representations before their inner product.

For position $t\in[L]$, let
\[
    \mathcal R_t\in\mathbb R^{d_H\times d_H}
\]
be the corresponding RoPE rotation matrix. By construction,
\[
    \mathcal R_t^\top\mathcal R_t=I_{d_H}.
\]
For a query position $l$ and a key position $j$, define the relative
rotation
\[
    \mathcal R_{l,j}
    :=
    \mathcal R_l^\top\mathcal R_j.
\]
Therefore,
\[
    \mathcal R_{l,j}^\top\mathcal R_{l,j}=I_{d_H},
    \qquad
    \|\mathcal R_{l,j}\|_2=1.
\]

The RoPE attention logit between positions $l$ and $j$ is
\[
\begin{aligned}
    z_{lj}
    &=
    (\mathcal R_l W_Qx_l)^\top
    (\mathcal R_j W_Kx_j)
    \\
    &=
    x_l^\top
    W_Q^\top
    \mathcal R_l^\top\mathcal R_j
    W_K
    x_j
    \\
    &=
    x_l^\top
    W_Q^\top
    \mathcal R_{l,j}
    W_K
    x_j.
\end{aligned}
\]

Thus, for each relative position, the effective interaction matrix is
\[
    B_{l,j}
    :=
    W_Q^\top\mathcal R_{l,j}W_K
    \in\mathbb R^{D\times D}.
\]

\begin{proposition}[Head-dimension reduction under RoPE]
\label{prop: rope}
Let $W_Q,W_K\in\mathbb R^{d_H\times D}$, and let
$W_{Q,r}$ and $W_{K,r}$ be their rank-$r$ truncated-SVD
approximations, where $r<d_H$. Define
\[
    \widetilde B_{l,j}
    :=
    W_{Q,r}^\top
    \mathcal R_{l,j}
    W_{K,r}.
\]
Then $\operatorname{rank}(\widetilde B_{l,j})\le r$, and uniformly
over all $l,j\in[L]$,
\[
\begin{aligned}
    \|B_{l,j}-\widetilde B_{l,j}\|_2
    \le\;
    \|W_Q\|_2\,\sigma_{r+1}(W_K)
    +
    \sigma_{r+1}(W_Q)\,\|W_K\|_2.
\end{aligned}
\]
Consequently, if $\|x_t\|_2\le B$ for all $t\in[L]$, then
\[
    |z_{lj}-\widetilde z_{lj}|
    \le
    B^2
    \left[
        \|W_Q\|_2\,\sigma_{r+1}(W_K)
        +
        \sigma_{r+1}(W_Q)\,\|W_K\|_2
    \right].
\]
\end{proposition}

\begin{proof}
Write
\[
    W_Q=W_{Q,r}+E_Q,
    \qquad
    W_K=W_{K,r}+E_K.
\]
By the Eckart--Young--Mirsky theorem,
\[
    \|E_Q\|_2=\sigma_{r+1}(W_Q),
    \qquad
    \|E_K\|_2=\sigma_{r+1}(W_K),
\]
and
\[
    \|W_{Q,r}\|_2=\|W_Q\|_2,
    \qquad
    \|W_{K,r}\|_2=\|W_K\|_2.
\]

For any pair $(l,j)$, we have
\[
\begin{aligned}
B_{l,j}-\widetilde B_{l,j}
&=
W_Q^\top\mathcal R_{l,j}W_K
-
W_{Q,r}^\top\mathcal R_{l,j}W_{K,r}
\\
&=
W_{Q,r}^\top\mathcal R_{l,j}E_K
+
E_Q^\top\mathcal R_{l,j}W_K.
\end{aligned}
\]
Hence,
\[
\begin{aligned}
\|B_{l,j}-\widetilde B_{l,j}\|_2
\le
\|W_{Q,r}\|_2
\|\mathcal R_{l,j}\|_2
\|E_K\|_2
+
\|E_Q\|_2
\|\mathcal R_{l,j}\|_2
\|W_K\|_2.
\end{aligned}
\]
Since $\mathcal R_{l,j}$ is orthogonal,
\[
    \|\mathcal R_{l,j}\|_2=1.
\]
Therefore,
\[
\begin{aligned}
\|B_{l,j}-\widetilde B_{l,j}\|_2
\le
\|W_Q\|_2\,\sigma_{r+1}(W_K)
+
\sigma_{r+1}(W_Q)\,\|W_K\|_2.
\end{aligned}
\]

Moreover,
\[
\begin{aligned}
|z_{lj}-\widetilde z_{lj}|
=
\left|
x_l^\top
(B_{l,j}-\widetilde B_{l,j})
x_j
\right|
\le
\|x_l\|_2
\|B_{l,j}-\widetilde B_{l,j}\|_2
\|x_j\|_2,
\end{aligned}
\]
which, together with $\|x_t\|_2\le B$, gives the stated logit
approximation bound.
\end{proof}

\vspace{1.cm}

\section{Approximation and tradeoff results for nonlinear target}\label{appendix: example}

\subsection{Tradeoff in Theorem~\ref{section: tradeoff}}\label{appendix: tradeoff}

\begin{figure}[htbp]
    \centering
    \begin{minipage}{0.5\textwidth}
        \centering
        \includegraphics[width=\textwidth]{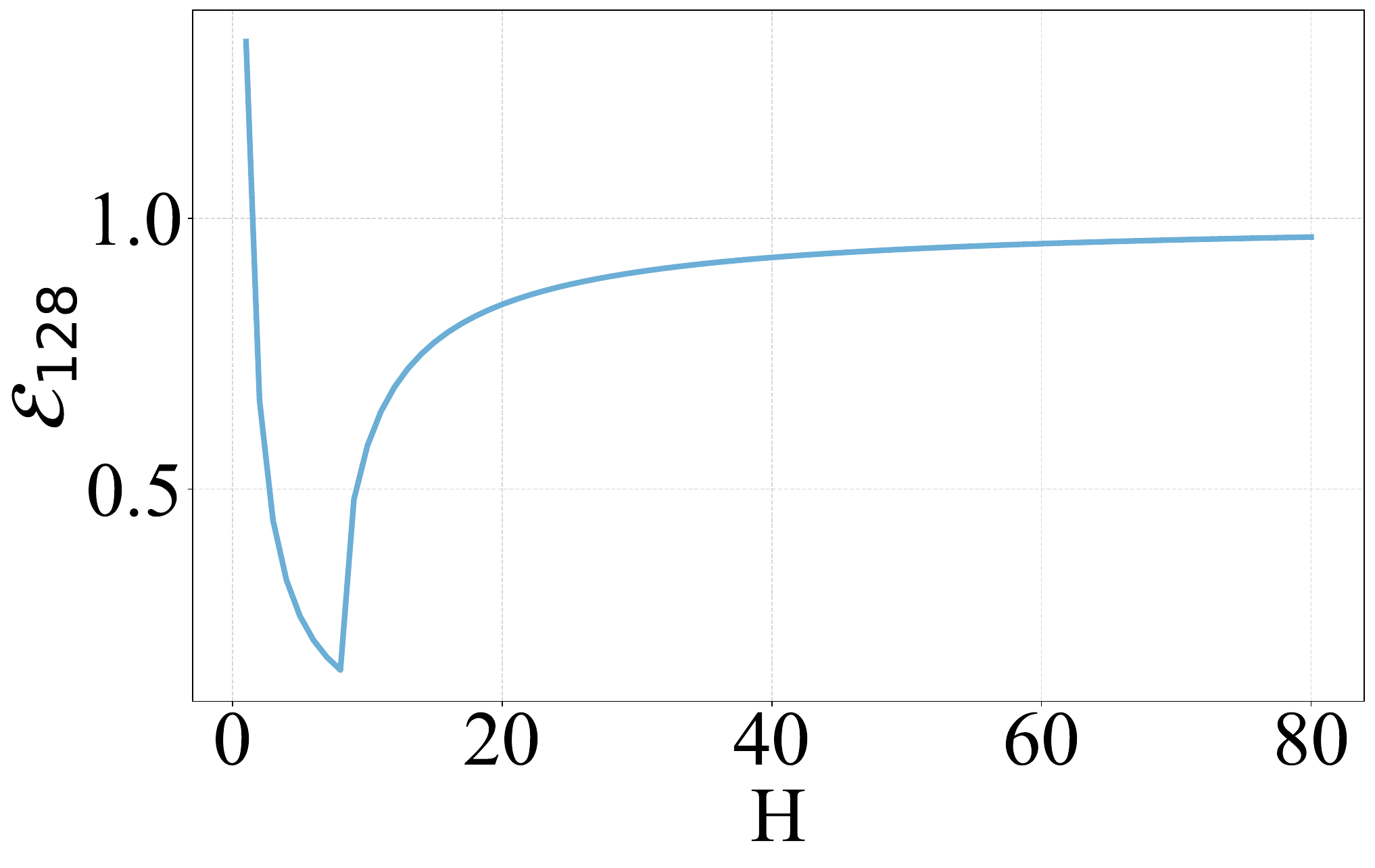}
        \caption{Induction Head Tradeoff}
        \label{exp: tradeoff plot}
    \end{minipage}
\end{figure}

We numerically validate the parameter trade-off in the approximation error estimate (i.e. the first equation of (\ref{equation: tradeoff opt})). Assume that all $d_m$'s are equal and there is no truncation error, we take $\rho=1,\ d=16$ and $D=128$. Then
$$\mathcal{E}_{128}=\sqrt{1-\frac{8}{H}}+\frac{1.3\ e^{0.02}}{H}$$
without concentration error.  We plot $\mathcal{E}(128)$ in Figure~\ref{exp: tradeoff plot}, which clearly shows a parameter trade-off trend. The trend of the continuous curve closely resembles that of the discrete polyline in Figure~\ref{fig: all-plots}.

\subsection{Further Experiments for Section \ref{section: tradeoff}}\label{appendix: further task}

\begin{figure}[htbp]
    \centering
    \begin{minipage}{0.45\textwidth}
        \centering
        \includegraphics[width=\textwidth]{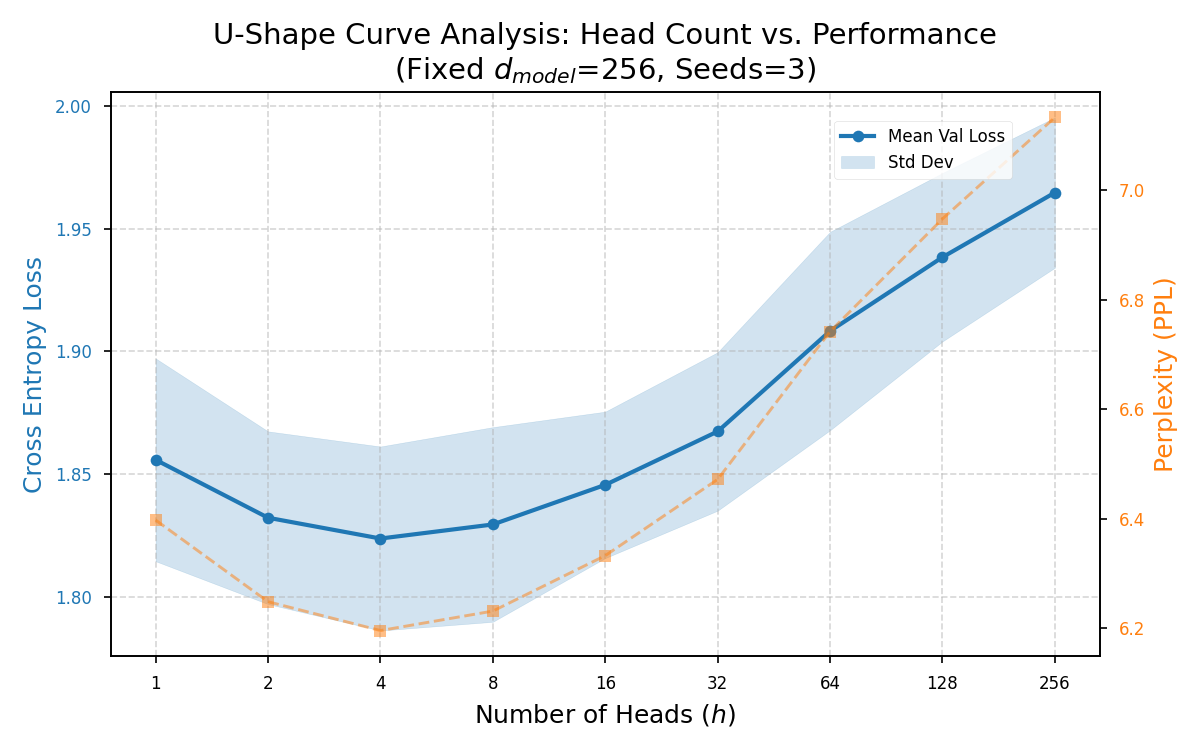}
        \caption{Next token prediction $D=256$.}
        \label{exp: ntp256}
    \end{minipage}
    \hfill
    \begin{minipage}{0.45\textwidth}
        \centering
        \includegraphics[width=\textwidth]{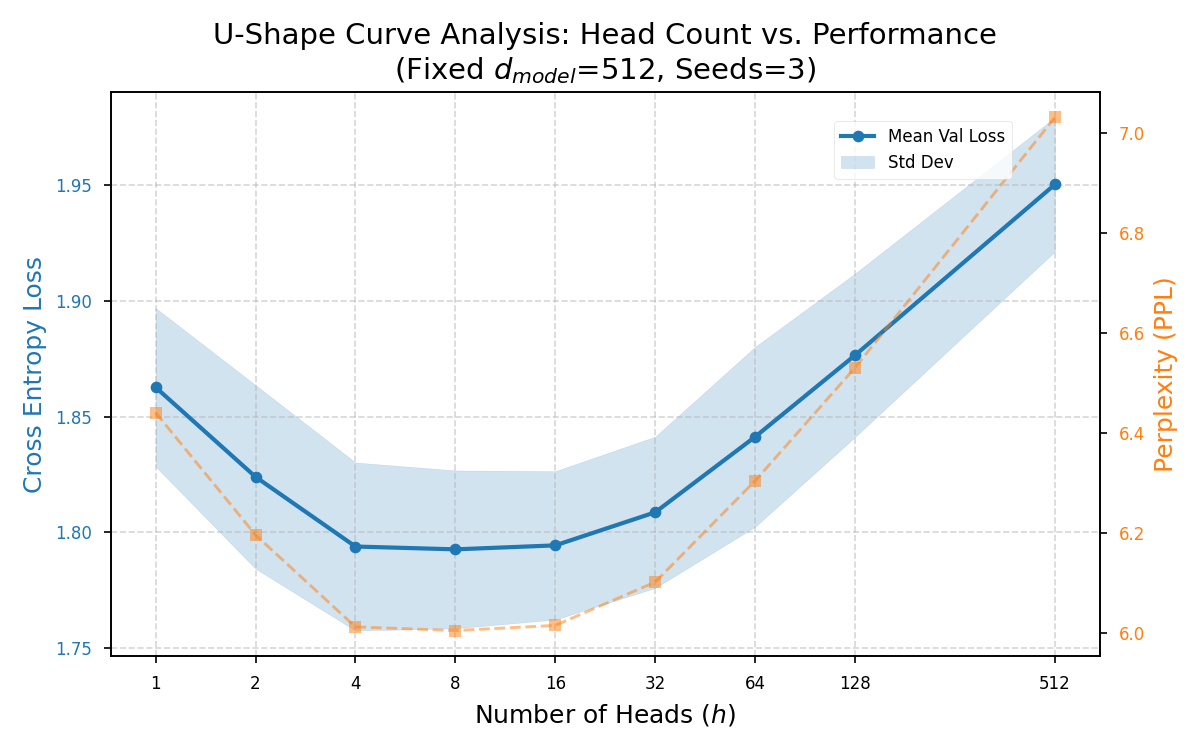}
        \caption{Next token prediction $D=512$.}
        \label{exp: ntp512}
    \end{minipage}
\end{figure}

\begin{figure}[htbp]
    \centering
    \begin{minipage}{0.45\textwidth}
        \centering
        \includegraphics[width=\textwidth]{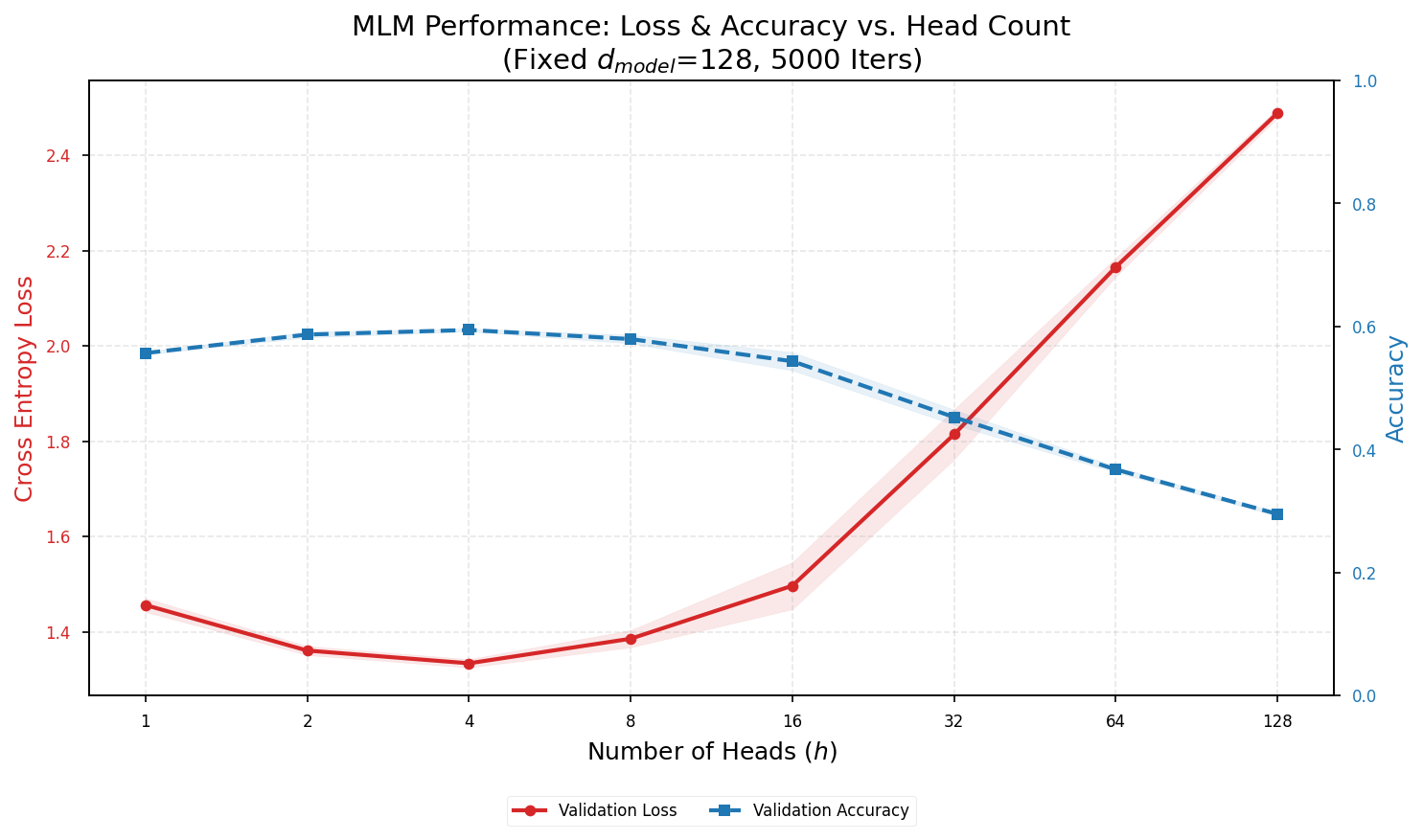}
        \caption{Masked language modeling $D=128$.}
        \label{exp: mlm128}
    \end{minipage}
    \hfill
    \begin{minipage}{0.45\textwidth}
        \centering
        \includegraphics[width=\textwidth]{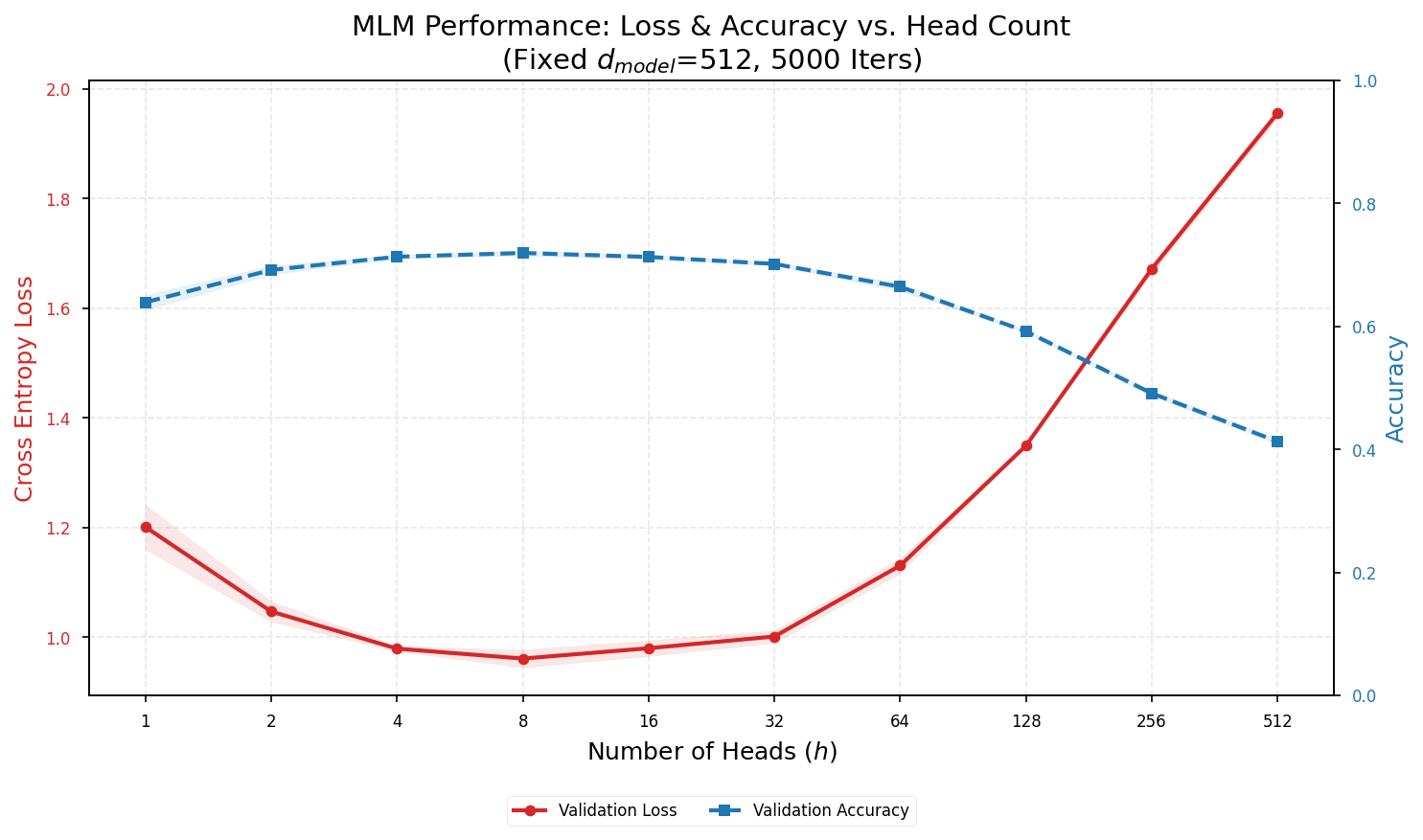}
        \caption{Masked language modeling $D=512$.}
        \label{exp: mlm512}
    \end{minipage}
\end{figure}

We first give out the empirical results for other values of $D$ for Figure \ref{fig: all-plots} in Figure \ref{exp: ntp256}, \ref{exp: ntp512}, \ref{exp: mlm128} and \ref{exp: mlm512}.

\begin{figure}[htbp]
    \centering
    \begin{minipage}{0.5\textwidth}
        \centering
        \includegraphics[width=\textwidth]{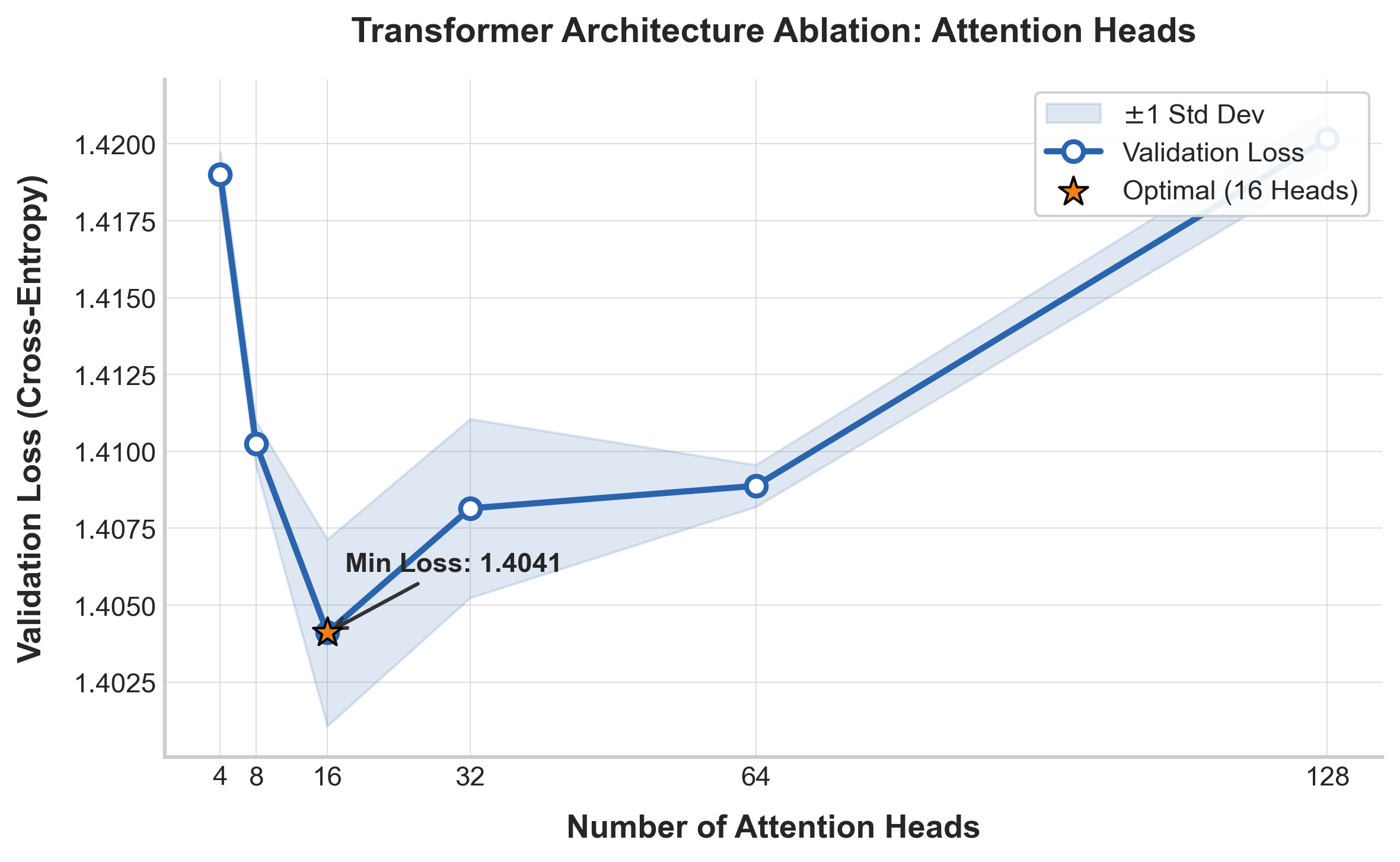}
        \caption{Induction Head Tradeoff}
        \label{exp: math}
    \end{minipage}
\end{figure}

To investigate the architectural scaling properties of multi-head attention (MHA), we conducted a controlled experiment in Figure \ref{exp: math} to determine the optimal balance between the number of attention heads ($H$) and the head dimension ($d_h$) for a fixed model capacity. 
We configured a series of Transformer-based models with approximately 330 million parameters, maintaining a constant hidden dimension ($d_{model} = 1024$) and depth ($L = 12$). We varied the head count $h \in \{4, 8, 16, 32, 64, 128\}$, which inversely corresponds to head dimensions $d_h \in \{256, 128, 64, 32, 16, 8\}$. 
The models were trained on a high-quality subset of the Stanford Alpaca dataset consisting of 15,000 samples, specifically filtered to emphasize logic and code-related reasoning. Each configuration was trained for 1,000 steps using a cosine learning rate scheduler (peak learning rate $2.5 \times 10^{-4}$) and a global batch size of 16. To ensure statistical robustness, we averaged the results across two distinct random seeds ($S \in \{42, 2026\}$). Generalization performance was measured via cross-entropy loss on a held-out validation set. This experimental setup allows us to identify the "Goldilocks" zone for attention granularity, where the model maximizes its representational capacity without suffering from the noise associated with excessively small head dimensions.

\subsection{Induction Head}\label{appendix: ih}

The former work ~\cite{wang2025transformersrichapproximationdynamics} defined a generalized induction head:
$$\mathrm{IH}_{n:t}(\boldsymbol{X})=\sum_{s=n}^{t-1} \pi_s\boldsymbol{x}_s,$$ 
where 
$$\pi_s=\mathrm{softmax}\bigg(\left(\boldsymbol{X}_{t-n+2:t}\boldsymbol{W}^*\boldsymbol{X}_{v-n+1:v-1}\right)_{v=n}^{L-1}\bigg)_{v=s},$$
and empirically pointed out that the first Transformer layer is used to extract information. In fact, the parameter allocation problem for the first layer reduces to the $n$-gram case.

\begin{figure}[htbp]
    \centering
    \begin{minipage}{0.5\textwidth}
        \centering
        \includegraphics[width=\textwidth]{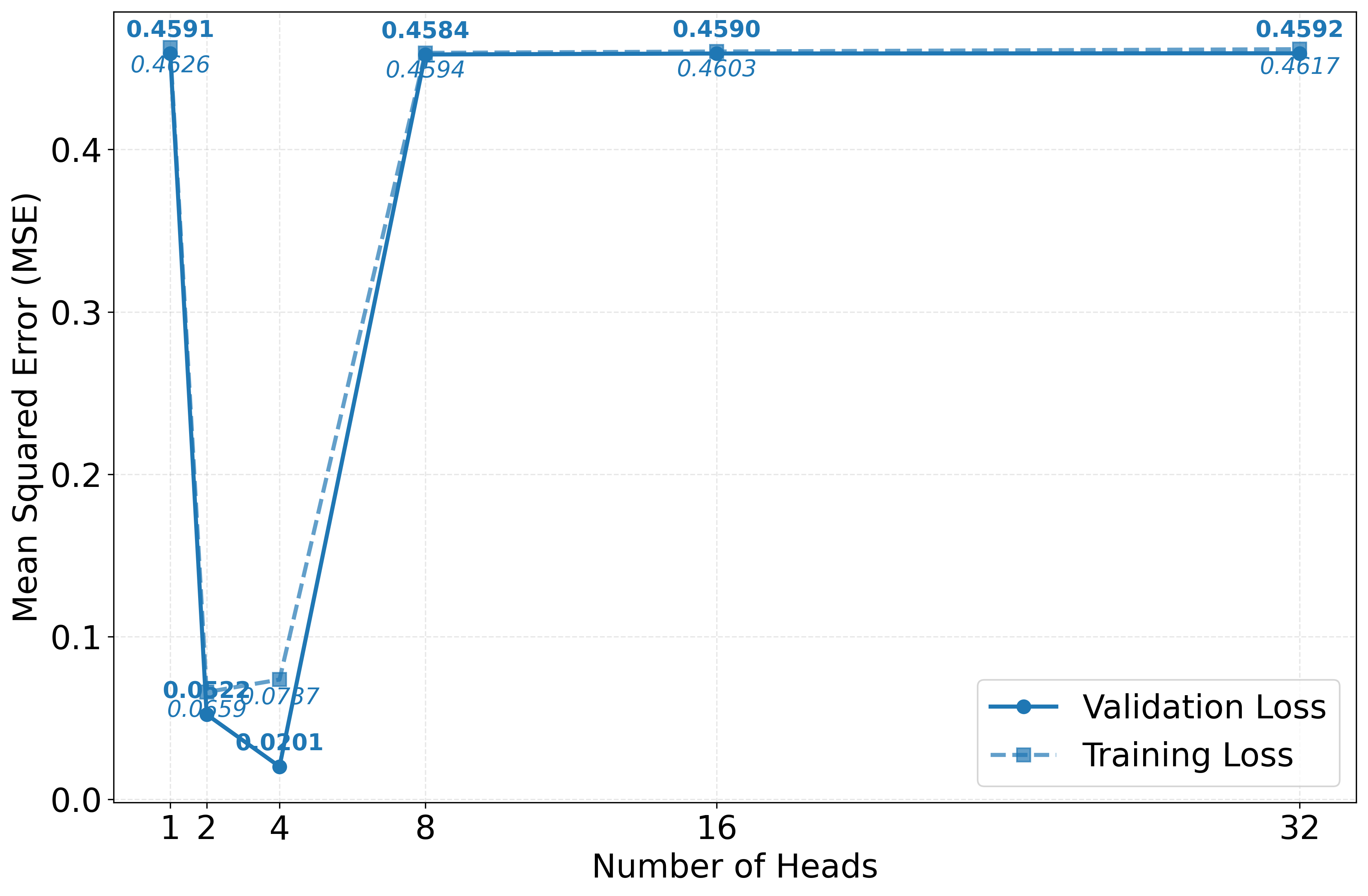}
        \caption{Induction Head Tradeoff}
        \label{exp: induction tradeoff}
    \end{minipage}
\end{figure}

We systematically investigated the effect of Transformer architectural variants on the Induction Head task  in Figure~\ref{exp: induction tradeoff}, which tests a model’s ability to recognize and extend repeated patterns—a fundamental reasoning mechanism for in-context learning. The task was constructed using synthetic sequences of length 128 with 10 unique tokens and embedded 5-token repeating patterns, yielding a dataset of 10,000 sequences (80\%/20\% train/validation).

The model architecture was a two-layer Transformer encoder with $d_{\mathrm{model}}=256$, pre-norm configuration, GELU feed-forward networks, learned positional encodings, and a two-layer MLP output head. We varied the number of attention heads in ${1,2,4,8,16,32}$ while keeping $d_{\mathrm{model}}$ fixed. Models were trained for 500 epochs using AdamW (learning rate $\in {10^{-3},5\times 10^{-4},10^{-4}}$, weight decay $10^{-4}$), cosine scheduling with ReduceLROnPlateau, gradient clipping (1.0), and mixed precision. Each configuration was repeated with three random seeds, and the best-performing model was selected by validation mean squared error (MSE).

Evaluation included validation MSE, heatmaps of performance across architectural variants, and qualitative prediction analysis. The results reveal how the number of heads influences the ability of the model to capture the Induction Head mechanism and demonstrate the robustness of certain configurations across seeds. This provides practical guidance for architectural design in tasks requiring in-context learning.

\subsection{Continuous Time-homogeneous System}\label{appendix: volterra}
In this subection, we analyze the approximation error of a continuous target function and derive the resulting trade-off.

We consider a sequence-to-sequence model in which both the input $\boldsymbol{X} = (\boldsymbol{x}_t)_{t\in\mathbb{Z}} \subset\mathcal{X}\subset \mathbb{R}^{d\times\mathbb{Z}}$ and output $\boldsymbol{Y} = (\boldsymbol{y}_t)_{t\in\mathbb{Z}} \subset\mathcal{Y}\subset \mathbb{R}^{d\times\mathbb{Z}}$ are sequences of infinite length and input tokens $\boldsymbol{x}_t$ are i.i.d. (we need a stronger assumption due to the target) random vectors with zero mean and identity covariance matrix and whose norm is bounded by $B$.
${H}$ denotes the mapping from $\mathcal{X}$ to $\mathcal{Y}$ with $\boldsymbol{y}_t = \mathrm{H}_t(\boldsymbol{X})$. 

To give the mapping $\mathrm{H}$ practical relevance, we must impose certain properties on the target function space. We now present the definitions of these properties.
\begin{definition}
$\mathrm{H}=\left\{\mathrm{H}_t:\mathcal{X}\mapsto\mathbb{R}^d;t\in \mathbb{R}\right\}$ be a sequence of functionals.

1.(\textbf{Linear}) $\mathrm{H}_t$ is linear if for any $\lambda,\lambda'\in\mathbb{R}$ and $\boldsymbol{X},\boldsymbol{X}'\in\mathcal{X}$, $\mathrm{H}_t(\lambda \boldsymbol{X}+\lambda'\boldsymbol{X}')=\lambda \mathrm{H}_t(\boldsymbol{X})+\lambda \mathrm{H}_t(\boldsymbol{X}')$.

2.(\textbf{Continuous}) $\mathrm{H}_t$ is continuous if for any $\boldsymbol{X},\boldsymbol{X}'\in\mathcal{X}$, $\lim_{\boldsymbol{X}'\rightarrow\boldsymbol{X}}|\mathrm{H}_t(\boldsymbol{X}')- \mathrm{H}_t(\boldsymbol{X})|=0$.

3.(\textbf{Bounded}) $\mathrm{H}_t$ is bounded if $\sup_{\boldsymbol{X}\in\mathcal{X},\boldsymbol{X}\neq\boldsymbol{0}}\frac{|\mathrm{H}_t(\boldsymbol{X})|}{\|\boldsymbol{X}\|_{\infty}}\leq\infty$.

4.(\textbf{Time-homogeneous}) $\mathrm{H}={\mathrm{H}_t:t\in\mathbb{R}}$ is time-homogeneous (or shift-equivariant) if the input-output relationship commutes with time shift: let $[S_{\tau}(\boldsymbol{X})]_t=\boldsymbol{x}_{t-\tau}$ be a shift operator, then $\mathrm{H}(S_{\tau}\boldsymbol{X})=S_{\tau}\mathrm{H}(\boldsymbol{X})$.

5.(\textbf{Causal}) $\mathrm{H}_t$ is causal if it does not depend on future values of the input. That is, if $\boldsymbol{X},\boldsymbol{X}'$ satisfy $\boldsymbol{x}_t=\boldsymbol{x}_t'$ for any $t\leq t_0$, then $\mathrm{H}_t(\boldsymbol{X})=\mathrm{H}_t(\boldsymbol{X}')$ for any $t\leq t_0$.

6.(\textbf{Regular}) $\mathrm{H}_t$ is regular if for any sequence ${\boldsymbol{X}^{(n)}\in \mathcal{X}:n\in\mathbb{Z}}$ such that $x^{(n)}_t\rightarrow \boldsymbol{0}$, then
$$\lim_{n\rightarrow\infty} \mathrm{H}_t(\boldsymbol{X}^{(n)})=\boldsymbol{0}.$$

\end{definition}

According to the following lemma, the target we studied in Section \ref{section: tradeoff} is a continuous, linear, regular, causal and time-homogeneous functional.

\begin{lemma}
Let $\mathrm{H}$ be a family of continuous, linear, regular, causal and time-homogeneous functionals on $\mathcal{X}$. Then, there exist a sequence $\rho:\mathbb{N}\rightarrow \mathbb{R}^d$ that is summable, i.e.
$$\|\rho\|_{l^1}:=\sum_{i=1}^d\sum_{j=0}^{\infty}|(\rho_i)_j|<\infty$$
and
$$\mathrm{H}_t(\boldsymbol{X})=\sum_{i=0}^{\infty}\rho_i\boldsymbol{x}_{t-i}, \ t\in\mathbb{Z}.$$
In particular, $H$ is uniformly bounded with $\sup_t\|H_t\|=\|\rho\|_{\ell^1}$.
\end{lemma}

\begin{proof}
We prove the existence of a summable sequence $\rho: \mathbb{N} \to \mathbb{R}^d$ satisfying the stated properties. Let $\mathcal{X} = \ell^\infty(\mathbb{Z}; \mathbb{R}^d)$ with the sup-norm $\|\boldsymbol{X}\| = \sup_{k \in \mathbb{Z}} \max_{1 \leq i \leq d} |X_{k,i}|$.

\vspace{0.5em}\noindent
\textbf{Step 1: Riesz representation and causality.} \\
Since each $\mathrm{H}_t: \mathcal{X} \to \mathbb{R}$ is continuous, linear, and regular, by the Riesz representation theorem for $\ell^\infty$ spaces with regular functionals, there exists a unique \textit{finitely additive} signed vector measure $\boldsymbol{\mu}_t$ on $\mathbb{Z}$ such that
\[
\mathrm{H}_t(\boldsymbol{X}) = \int_{\mathbb{Z}} \boldsymbol{X}^T d\boldsymbol{\mu}_t = \sum_{k \in \mathbb{Z}} \boldsymbol{X}_k^T \boldsymbol{\mu}_t(\{k\}),
\]
where $\boldsymbol{\mu}_t(\{k\}) \in \mathbb{R}^d$ is the mass at integer $k$, and $\|H_t\| = \|\boldsymbol{\mu}_t\| = \sum_{k \in \mathbb{Z}} \sum_{i=1}^d |\mu_{t,i}(\{k\})|$. 

Causality implies that $\mathrm{H}_t(\boldsymbol{X})$ depends only on $\{\boldsymbol{X}_k : k \leq t\}$. Hence, $\boldsymbol{\mu}_t(\{k\}) = \mathbf{0}$ for all $k > t$, and the representation simplifies to:
\begin{equation}
\mathrm{H}_t(\boldsymbol{X}) = \sum_{k=-\infty}^t \boldsymbol{X}_k^T \boldsymbol{\mu}_t(\{k\}). \tag{1}
\end{equation}

\vspace{0.5em}\noindent
\textbf{Step 2: Time homogeneity and shift invariance.} \\
For any $\tau \in \mathbb{Z}$, time homogeneity implies $\mathrm{H}_t(\boldsymbol{X}) = \mathrm{H}_{t+\tau}(\boldsymbol{X}^{(\tau)})$ where $\boldsymbol{X}^{(\tau)}_k = \boldsymbol{X}_{k-\tau}$. Applying (1) to the right-hand side:
\[
\mathrm{H}_{t+\tau}(\boldsymbol{X}^{(\tau)}) = \sum_{m=-\infty}^{t+\tau} \boldsymbol{X}_{m-\tau}^T \boldsymbol{\mu}_{t+\tau}(\{m\}).
\]
Substituting $j = m - \tau$:
\[
\mathrm{H}_{t+\tau}(\boldsymbol{X}^{(\tau)}) = \sum_{j=-\infty}^t \boldsymbol{X}_j^T \boldsymbol{\mu}_{t+\tau}(\{j + \tau\}).
\]
Equating with (1) and comparing coefficients:
\begin{equation}
\boldsymbol{\mu}_t(\{j\}) = \boldsymbol{\mu}_{t+\tau}(\{j + \tau\}) \quad \forall j \leq t. \tag{2}
\end{equation}

\vspace{0.5em}\noindent
\textbf{Step 3: Construction of $\rho$.} \\
Fix $t = 0$ in (2) and set $\tau = -t$:
\[
\boldsymbol{\mu}_t(\{j\}) = \boldsymbol{\mu}_{0}(\{j + t\}) \quad \forall j \leq t.
\]
Define $\boldsymbol{\mu}(\{k\}) = \boldsymbol{\mu}_{0}(\{k\})$. By causality, $\boldsymbol{\mu}(\{k\}) = \mathbf{0}$ for $k > 0$. Using time homogeneity:
\[
\mathrm{H}_t(\boldsymbol{X}) = \mathrm{H}_0(\boldsymbol{X}^{(-t)}) = \sum_{k=-\infty}^{0} \boldsymbol{X}_{k+t}^T \boldsymbol{\mu}(\{k\}).
\]
Substituting $i = -k$:
\begin{equation}
\mathrm{H}_t(\boldsymbol{X}) = \sum_{i=0}^{\infty} \boldsymbol{X}_{t - i}^T \boldsymbol{\mu}(\{-i\}). \tag{3}
\end{equation}
Define $\rho_i = \boldsymbol{\mu}(\{-i\})$ for $i \geq 0$. Then (3) becomes:
\[
\mathrm{H}_t(\boldsymbol{X}) = \sum_{i=0}^{\infty} \boldsymbol{x}_{t-i}^T \rho_i.
\]

\vspace{0.5em}\noindent
\textbf{Step 4: Summability and norm equality.} \\
From the Riesz representation and causality:
\[
\|\rho\|_{\ell^1} = \sum_{i=0}^{\infty} \sum_{j=1}^d |(\rho_i)_j| = \sum_{k=-\infty}^{0} \sum_{j=1}^d |\mu_j(\{k\})| = \|\boldsymbol{\mu}\| = \|H_0\| < \infty.
\]
For any $t \in \mathbb{Z}$, using (2) and shift invariance:
\[
\|\mathrm{H}_t\| = \sum_{j=-\infty}^t \sum_{i=1}^d |\mu_{t,i}(\{j\})| = \sum_{j=-\infty}^t \sum_{i=1}^d |\mu_i(\{j + t\})| = \|\boldsymbol{\mu}\| = \|\rho\|_{\ell^1}.
\]
Thus $\sup_{t \in \mathbb{Z}} \|\mathrm{H}_t\| = \|\rho\|_{\ell^1}$.
\end{proof}

We now consider a continuous time-homogeneous system. According to Theorem C.6 in \cite{wang2024stablessmalleviatingcursememory} and \cite{1456575}, we obtain the following decomposition result for such systems:

\begin{lemma}[Volterra Series Decomposition]
$\mathrm{H}$ is a continuous time-homogeneous system with input $\boldsymbol{X}$ and output $\boldsymbol{Y}$, then $\mathrm{H}$ can be expanded in the Volterra series as follows
\begin{equation}
    \boldsymbol{y}_t=\mathrm{H}_t(\boldsymbol{X})=h_0+\sum_{n=1}^{\infty} \sum_{\tau_1=0}^{\infty}\cdots\sum_{\tau_n=0}^{\infty} \mathrm{H}^{(n)}(\tau_1,\ldots,\tau_n)\big(\boldsymbol{x}_{t-\tau_1}\otimes\cdots\otimes\boldsymbol{x}_{t-\tau_n}\big),
\end{equation}
where $\otimes$ denotes the Kronecker product. In particular, we call the expansion order $n$ to be the series' order.
\end{lemma}

We focus on the term $\mathrm{H}^{(n)}$ and derive its approximation error when using a single-layer Transformer to approximate it. We begin by stating the lemmas required for our analysis:

\begin{lemma}
    Let $\{x_i\}_{i=1}^n$ be i.i.d.\ random variables with $\mathbb{E}[x_i] = 0$ and $\mathbb{E}[x_i^2] = 1$. 
    Then $\bigotimes_{i=1}^{n} x_i$ is a random vector with mean zero and identity covariance matrix.
\end{lemma}

\begin{proof}
    It suffices to consider the scalar case $d=1$. Define $y := \prod_{i=1}^n x_i$. Then
    $$
        \mathbb{E}[y] = \mathbb{E}\left[\prod_{i=1}^n x_i\right] = \prod_{i=1}^n \mathbb{E}[x_i] = 0,
    $$
    and
    $$
        \mathbb{E}[y^2] = \mathbb{E}\left[\left(\prod_{i=1}^n x_i\right)^2\right] 
        = \prod_{i=1}^n \mathbb{E}[x_i^2] = 1.
    $$
    Thus $\mathrm{Var}(y) = 1$, which proves the claim.
\end{proof}

\begin{lemma}
    $\boldsymbol{x}_i$ are random vectors with $\mathbb{E}\|\boldsymbol{x}_i\|_2=B_i$, and $\|\boldsymbol{x}_i-\bar{\boldsymbol{x}}_i\|_2\leq\varepsilon_i\leq\varepsilon$. Then $\mathbb{E}\|\bigotimes_{i=1}^{n} \tilde{\boldsymbol{x}}_i-\bigotimes_{i=1}^{n} \hat{\boldsymbol{x}}_i\|_2\leq\prod_{i=1}^n(B_i+\varepsilon_i)-\prod_{i=1}^n B_i$
\end{lemma}

\begin{proof}
    From the property of projection operation,
    $$\|\tilde{\boldsymbol{x}}_i-\hat{\boldsymbol{x}}_i\|_2\leq\|P\|_2\|\boldsymbol{x}_i-\hat{\boldsymbol{x}}_i\|_2\leq\varepsilon_i\leq\varepsilon.$$
    By direct calculation,
    \begin{equation*}
        \begin{aligned}
            \mathbb{E}\|\bigotimes_{i=1}^{n} \tilde{\boldsymbol{x}}_i-\bigotimes_{i=1}^{n} \hat{\boldsymbol{x}}_i\|_2&\leq\sum_{\phi\neq S\subset\{1,\ldots,n\}}\left(\prod_{i\in S}\varepsilon_i\right)\mathbb{E}\|\bigotimes_{j\notin S}\tilde{\boldsymbol{x}}_j\|_2\\
            &=\sum_{\phi\neq S\subset\{1,\ldots,n\}}\left(\prod_{i\in S}\varepsilon_i\right)\prod_{j\notin S}B_j\\
            &=\prod_{i=1}^n(B_i+\varepsilon_i)-\prod_{i=1}^n B_i.
        \end{aligned}
    \end{equation*}
\end{proof}


We use the FFN to implement the Kronecker product in our proof; therefore, the universal approximation property of FFNs is also required.

\begin{definition}[Barron space~(\cite{e2019apriori,weinan2021barron,ma2020towards})]
Consider functions $f:X\to\mathbb{R}$ that admit the following representation:
$f(\boldsymbol{x})=\int_{\Omega}a\sigma(\boldsymbol{b}^\top\boldsymbol{x}+c)\rho(\mathrm{d} a,\mathrm{d} \boldsymbol{b},\mathrm{d} c),\ \boldsymbol{x}\in X$.
For any $p\in[1,+\infty]$, we define the Barron norm as $\|{f}\|_{\mathcal{B}_p}:=\inf_{\rho}\Big    (\mathbb{E}_{\rho}\left[|a|^p(\|{\boldsymbol{b}}
\|_1+|c|)^p\right]\Big)^{1/p}$.
Then the Barron space are defined as:
$\mathcal{B}_p:=\{f\in\mathcal{C}:\|{f}\|_{\mathcal{B}_p}<+\infty\}$.
\end{definition}

\begin{lemma}[\cite{ma2020towards}]\label{lemma barron}
    For any $f\in\mathcal{B}$, there exists a two-layer ReLU neural network $\mathrm{FFN}(\boldsymbol{x})=\sum\limits_{k=1}^K a_k\sigma(\boldsymbol{b}_k^\top\boldsymbol{x}+c_k)$ with $M$ neurons such that
    \begin{align*}
        \|{f-\mathrm{FFN}}\|_{L^\infty[-2,2]}\leq \mathcal{O}\left(\frac{\|{f}\|_{\mathcal{B}}\sqrt{\log K}}{\sqrt{K}}\right).
    \end{align*}
\end{lemma}

We now present the approximation and tradeoff result:

\begin{theorem}[Tradeoff for Nonlinear Target]
\label{theorem: nonlinear tradeoff}
Consider the $n$-th order Volterra term
\[
\mathrm{H}_t(\boldsymbol{X})
=
\sum_{\tau_1=0}^{\infty}\cdots
\sum_{\tau_n=0}^{\infty}
\mathrm{H}^{(n)}(\tau_1,\ldots,\tau_n)
\big(
\boldsymbol{x}_{t-\tau_1}\otimes\cdots\otimes
\boldsymbol{x}_{t-\tau_n}
\big).
\]
Suppose that we have $H_m$ heads of dimension $d_m\le d$ for
$m=1,\ldots,M$, and that the model dimension
\[
D=\sum_{m=1}^{M}H_m d_m
\]
is fixed.

For $m\in[M]$, define
\[
\gamma_m
:=
\mathbb{I}_{\{d_m<d\}}
\left(
\sqrt{1-\frac{d_m}{d}}
+
C\sqrt{\frac{\log(2M/\delta)}{d}}
\right),
\]
and
\[
\eta_m
:=
B\frac{1.3e^{0.02m}}{H_m}.
\]
For the current token, which is preserved by the residual connection, set
\[
\gamma_0=\eta_0=0.
\]

Then, with probability at least $1-\delta$,
\begin{equation}
\label{equation: volterra}
\begin{aligned}
\mathcal{E}_D(\boldsymbol{X})
\leq&
\sum_{\tau_1=0}^{M}\cdots\sum_{\tau_n=0}^{M}
\big\|
\mathrm{H}^{(n)}(\tau_1,\ldots,\tau_n)
\big\|_2
\Bigg[
B^n\sum_{j=1}^{n}\gamma_{\tau_j}
+
\prod_{j=1}^{n}
\left(B+\eta_{\tau_j}\right)
-B^n
\Bigg]
+
\varepsilon_{\mathrm{H}}
+
\varepsilon_{\mathrm{FFN}}.
\end{aligned}
\end{equation}
Here,
\begin{equation}
\label{equation: epsilon H}
\begin{aligned}
\varepsilon_{\mathrm{H}}
:=
\sup_t
\Bigg\|
&
\sum_{\tau_1=0}^{\infty}\cdots
\sum_{\tau_n=0}^{\infty}
\mathrm{H}^{(n)}(\tau_1,\ldots,\tau_n)
\big(
\boldsymbol{x}_{t-\tau_1}\otimes\cdots\otimes
\boldsymbol{x}_{t-\tau_n}
\big)
\\
&-
\sum_{\tau_1=0}^{M}\cdots
\sum_{\tau_n=0}^{M}
\mathrm{H}^{(n)}(\tau_1,\ldots,\tau_n)
\big(
\boldsymbol{x}_{t-\tau_1}\otimes\cdots\otimes
\boldsymbol{x}_{t-\tau_n}
\big)
\Bigg\|_2,
\end{aligned}
\end{equation}
and $\varepsilon_{\mathrm{FFN}}$ denotes the approximation error
incurred when the FFN implements the required Kronecker-product
operations.
\end{theorem}

\begin{corollary}[Parameter allocation via trade-offs]
\label{corollary: allocation}
Under the same conditions as
Theorem~\ref{theorem: nonlinear tradeoff},
a principled parameter allocation can be obtained by solving
\begin{equation}
\label{equation: opt}
\begin{aligned}
\min_{M,\{H_m,d_m\}_{m=1}^{M}}
\quad&
\sum_{\tau_1=0}^{M}\cdots\sum_{\tau_n=0}^{M}
\big\|
\mathrm{H}^{(n)}(\tau_1,\ldots,\tau_n)
\big\|_2
\Bigg[
B^n\sum_{j=1}^{n}\gamma_{\tau_j}
\\
&\qquad\qquad\qquad
+
\prod_{j=1}^{n}
\left(B+\eta_{\tau_j}\right)
-B^n
\Bigg]
+
\varepsilon_{\mathrm{H}}(M)
\\
\mathrm{s.t.}\quad&
\sum_{m=1}^{M}H_m d_m=D,
\\
&
M,H_m,d_m\in\mathbb{N}_+,
\qquad
d_m\le d,
\end{aligned}
\end{equation}
where
\[
\gamma_m
=
\mathbb{I}_{\{d_m<d\}}
\left(
\sqrt{1-\frac{d_m}{d}}
+
C\sqrt{\frac{\log(2M/\delta)}{d}}
\right),
\qquad
\eta_m
=
B\frac{1.3e^{0.02m}}{H_m},
\]
with $\gamma_0=\eta_0=0$.
If $\varepsilon_{\mathrm{FFN}}$ is fixed independently of the allocation,
it can be omitted from the optimization objective.
\end{corollary}



\begin{proof}
The construction of the embedding layer and the multi-head attention layer
follows the same procedure as in the proof of Theorem~\ref{section: tradeoff}.
After these two steps, for each token we obtain
\[
\begin{pmatrix}
\widetilde{\boldsymbol{x}}_t\\
0\\
\vdots\\
0
\end{pmatrix}
\longrightarrow
\begin{pmatrix}
\widetilde{\boldsymbol{x}}_t\\
\widetilde{\boldsymbol{x}}_{t-1}\\
\vdots\\
\widetilde{\boldsymbol{x}}_{t-M}\\
0
\end{pmatrix}
:=\boldsymbol{x}^{(1/2)}_t .
\]

For each $m\in[M]$, let
$\boldsymbol{P}_m\in\mathbb{R}^{d_m\times d}$ be the random projection
used in the proof of Theorem~\ref{section: tradeoff}, and define
$\boldsymbol{\Pi}_m=\boldsymbol{P}_m^\top\boldsymbol{P}_m$.
By Lemma~\ref{lemma: prob projection} and a union bound over the $M$
projection groups, with probability at least $1-\delta$,
\[
\|(\boldsymbol{I}-\boldsymbol{\Pi}_m)\boldsymbol{x}\|_2
\le
B\left[
\mathbb{I}_{\{d_m<d\}}
\sqrt{1-\frac{d_m}{d}}
+
C\sqrt{\frac{\log(2M/\delta)}{d}}
\right],
\qquad m\in[M].
\]
Define
\[
\gamma_\delta
:=
\mathbb{I}_{\{d_{\min}<d\}}
\sqrt{1-\frac{d_{\min}}{d}}
+
C\sqrt{\frac{\log(2M/\delta)}{d}},
\qquad
d_{\min}:=\min_{m\in[M]}d_m .
\]
Then, for every $m\in[M]$,
\[
\|(\boldsymbol{I}-\boldsymbol{\Pi}_m)\boldsymbol{x}\|_2
\le B\gamma_\delta .
\]

For any $(\tau_1,\ldots,\tau_n)\in\{0,\ldots,M\}^n$, the difference
between the original and projected tensor representations admits the
telescoping decomposition
\[
\begin{aligned}
&
\boldsymbol{x}_{t-\tau_1}\otimes\cdots\otimes
\boldsymbol{x}_{t-\tau_n}
-
\widetilde{\boldsymbol{x}}_{t-\tau_1}\otimes\cdots\otimes
\widetilde{\boldsymbol{x}}_{t-\tau_n}
\\
&=
\sum_{j=1}^{n}
\widetilde{\boldsymbol{x}}_{t-\tau_1}\otimes\cdots\otimes
\widetilde{\boldsymbol{x}}_{t-\tau_{j-1}}
\otimes
\left(
\boldsymbol{x}_{t-\tau_j}
-
\widetilde{\boldsymbol{x}}_{t-\tau_j}
\right)
\otimes
\boldsymbol{x}_{t-\tau_{j+1}}\otimes\cdots\otimes
\boldsymbol{x}_{t-\tau_n}.
\end{aligned}
\]
Since $\|\boldsymbol{x}_t\|_2\le B$ and orthogonal projection is
non-expansive,
$\|\widetilde{\boldsymbol{x}}_t\|_2\le B$. Therefore,
\[
\left\|
\boldsymbol{x}_{t-\tau_1}\otimes\cdots\otimes
\boldsymbol{x}_{t-\tau_n}
-
\widetilde{\boldsymbol{x}}_{t-\tau_1}\otimes\cdots\otimes
\widetilde{\boldsymbol{x}}_{t-\tau_n}
\right\|_2
\le
nB^n\gamma_\delta .
\]

Furthermore, by Lemma~\ref{lemma: tensor error},
\[
\left\|
\widetilde{\boldsymbol{x}}_{t-\tau_1}\otimes\cdots\otimes
\widetilde{\boldsymbol{x}}_{t-\tau_n}
-
\widehat{\boldsymbol{x}}_{t-\tau_1}\otimes\cdots\otimes
\widehat{\boldsymbol{x}}_{t-\tau_n}
\right\|_2
\le
(B+\varepsilon_{\mathrm{Attn}})^n-B^n,
\]
where
\[
\varepsilon_{\mathrm{Attn}}
=
\max_{m\in[M]}
\frac{1.3e^{0.02m}}{H_m}.
\]

Combining the projection error, attention-extraction error, kernel
truncation error $\varepsilon_H$, and FFN approximation error
$\varepsilon_{\mathrm{FFN}}$, with probability at least $1-\delta$,
\[
\begin{aligned}
\mathcal{E}_D(\boldsymbol{X})
\le&
\sum_{\tau_1=0}^{M}\cdots\sum_{\tau_n=0}^{M}
\left\|
\boldsymbol{H}^{(n)}(\tau_1,\ldots,\tau_n)
\right\|_2
\left[
nB^n\gamma_\delta
+
(B+\varepsilon_{\mathrm{Attn}})^n-B^n
\right]
\\
&\quad
+\varepsilon_H+\varepsilon_{\mathrm{FFN}} .
\end{aligned}
\]
\end{proof}

\vspace{1.cm}

\section{Additional Experiments}\label{appendix: experiment}

\subsection{Inspiration of Parameter Allocation}

\begin{figure}[htbp]
    \centering
    \includegraphics[width=0.8\textwidth]{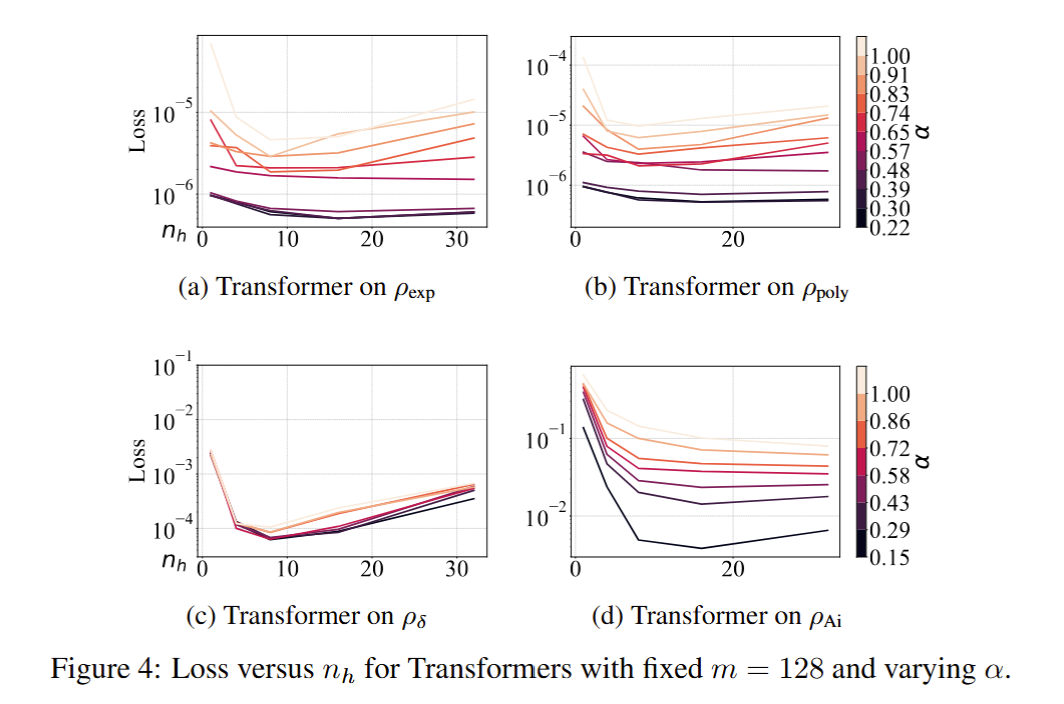}
        \caption{Linear Convolution Tradeoff}
        \label{exp: linear tradeoff}
\end{figure}

The former work \cite{jiang2025numericalinvestigationsequencemodeling} empirically investigated how the head trade-off is influenced by the memory structure and the difficulty of the target relationship in Figure \ref{exp: linear tradeoff}. As illustrated in Figure 4 of their work, the trade-off depends on both the underlying memory structure and the level of task difficulty. Each subfigure corresponds to a specific memory structure, with the plots showing loss as a function of the number of heads. The color of the curves encodes the parameter $\alpha$, which reflects task difficulty.
The results indicate that, for certain tasks, the trade-off remains consistent across different levels of difficulty (Figure (c)). In tasks (a) and (b), when the task is relatively easy, no clear trade-off is observed. However, in task (c), a trade-off appears consistently, regardless of task difficulty.

\subsection{Verification of Theorem \ref{theorem: softmax} and Corollary \ref{corollary: softmax}}

To verify Theorem \ref{theorem: softmax}: We rerun Figure~\ref{exp: jacobian spectral} for a complementary evaluation protocol. In the original manuscript, we varied the sequence length $L$ and computed the Jacobian (spectral) norm associated with the last query. Here, under the same model and data setting, we instead fix $L=1024$ and compute the Jacobian (spectral) norm associated with the $l$-th query (averaged over 10 sampled sequences) as follows. It is clearly to have $l\lVert J_l\rVert _2\in[0.9, 1.7]$ across reported positions, verifying the $\mathcal O(1/l)$-bound. 

\begin{table}[t]
\centering
\caption{Jacobian spectral norm at different query positions.}
\label{tab:jacobian_position}
\begin{tabular}{ccc}
\toprule
Query position $l\in[L]$
& Mean Jacobian spectral norm $\lVert J_l\rVert_2$
& $l\lVert J_l\rVert_2$ \\
\midrule
2    & $4.512\times 10^{-1}$ & $0.9024$ \\
4    & $2.792\times 10^{-1}$ & $1.1168$ \\
8    & $1.506\times 10^{-1}$ & $1.2048$ \\
16   & $7.413\times 10^{-2}$ & $1.1861$ \\
32   & $3.582\times 10^{-2}$ & $1.1462$ \\
64   & $1.753\times 10^{-2}$ & $1.1219$ \\
128  & $1.122\times 10^{-2}$ & $1.4362$ \\
256  & $5.718\times 10^{-3}$ & $1.4638$ \\
512  & $3.312\times 10^{-3}$ & $1.6957$ \\
1024 & $1.650\times 10^{-3}$ & $1.6896$ \\
\bottomrule
\end{tabular}
\end{table}

Consistency between Corollary \ref{corollary: softmax} and Figure 3: An implication of Corollary \ref{corollary: softmax} is, for most of queries at larger positions (say $l\ge 0.1L$), the distillation error per query scales like $\mathcal O(1/L)\to 0$ ($L\gg 1$); while for a small portion of queries at smaller positions (say $l\le 0.1L$), the corresponding Jacobian mapping is highly degenerated ($\mathrm{rank}(J_l)\le l\le 0.1L$) and hence controls the distillation error incurred by truncated SVD. This makes the SVD initialization a warm start, but also need further refinements to reduce the distillation error corresponding to earlier queries. Both steps together form Algorithm~\ref{algorithm: reduction}, which is implemented and verified in Figure~\ref{exp: trained L scale}. 

\subsection{Role of Layers in Figure~\ref{exp: low rank}}

We further evaluate layer-wise head-dimension reduction on
\textsc{LLaMA}-3-8B, measured by perplexity (PPL), by separately
compressing the deepest layers (21--24) and several early layers
(2, 3, and 4). The results are reported in
Table~\ref{tab:llama8b_layerwise}.

\begin{table}[t]
\centering
\caption{
Layer-wise head-dimension reduction on \textsc{LLaMA}-3-8B, evaluated by perplexity (PPL).
}
\label{tab:llama8b_layerwise}
\begin{tabular}{ccccc}
\toprule
Reduced rank $d_h$
& Deepest layers 21--24
& Layer 2
& Layer 3
& Layer 4 \\
\midrule
1   & 4.72 & 6.02 & 4.69 & 4.75 \\
4   & 4.72 & 5.61 & 4.61 & 4.76 \\
8   & 4.71 & 5.41 & 4.59 & 4.67 \\
16  & 4.68 & 5.22 & 4.57 & 4.58 \\
32  & 4.61 & 4.88 & 4.52 & 4.52 \\
64  & 4.51 & 4.57 & 4.45 & 4.47 \\
128 & 4.41 & 4.41 & 4.41 & 4.41 \\
\bottomrule
\end{tabular}
\end{table}

These results confirm a clear layer dependence of model reduction. Compressing the deepest layers causes almost no perplexity degradation even at very small ranks. Layers 3 and 4 are similarly robust, whereas layer 2 is moderately more sensitive. We will add these new experiments in the revision to emphasize that even the saturation behavior is layer-dependent, it is particularly pronounced in deep layers and hence leads to lossless later-layer model reduction. 

\subsection{Reducing Budget on Number of Heads}

We conducted a systematic investigation of Transformer architectural hyperparameters on the WikiText-103 language modeling task. The goal was to understand how attention head dimension and number of heads affect model performance in next-token prediction.

The models were 6-layer GPT-style Transformer decoders with learned token and positional embeddings (sequence length 256), RMSNorm, SwiGLU feed-forward networks, residual connections, and weight-tied output projections. Two hyperparameter sweeps were performed: (1) varying head dimension with 8 attention heads fixed, and (2) varying the number of heads (1–32) with head dimension fixed at 64. The model dimension $d_{\mathrm{model}}$ was set as the product of number of heads and head dimension.

Training was performed on $1\%$ of WikiText-103 (~1M tokens) for 2 epochs using AdamW (learning rate $3\times10^{-4}$, weight decay 0.1) with cosine scheduling, 200-step warmup, gradient clipping, dropout 0.1, and mixed precision. Models were evaluated on validation and test sets using cross-entropy loss and perplexity. Each configuration recorded training curves, validation losses, and test performance for analysis.

Comparative analysis reveals that cutting head dimension is a more effective way to reduce model parameters than reducing the number of heads, which incurs higher loss. The study highlights how architectural choices influence performance and parameter efficiency in language modeling, offering insights for balancing model capacity and computational budget.

\begin{figure}[htbp]
    \centering

    \begin{minipage}{0.50\textwidth}
        \centering
        \includegraphics[width=\textwidth]{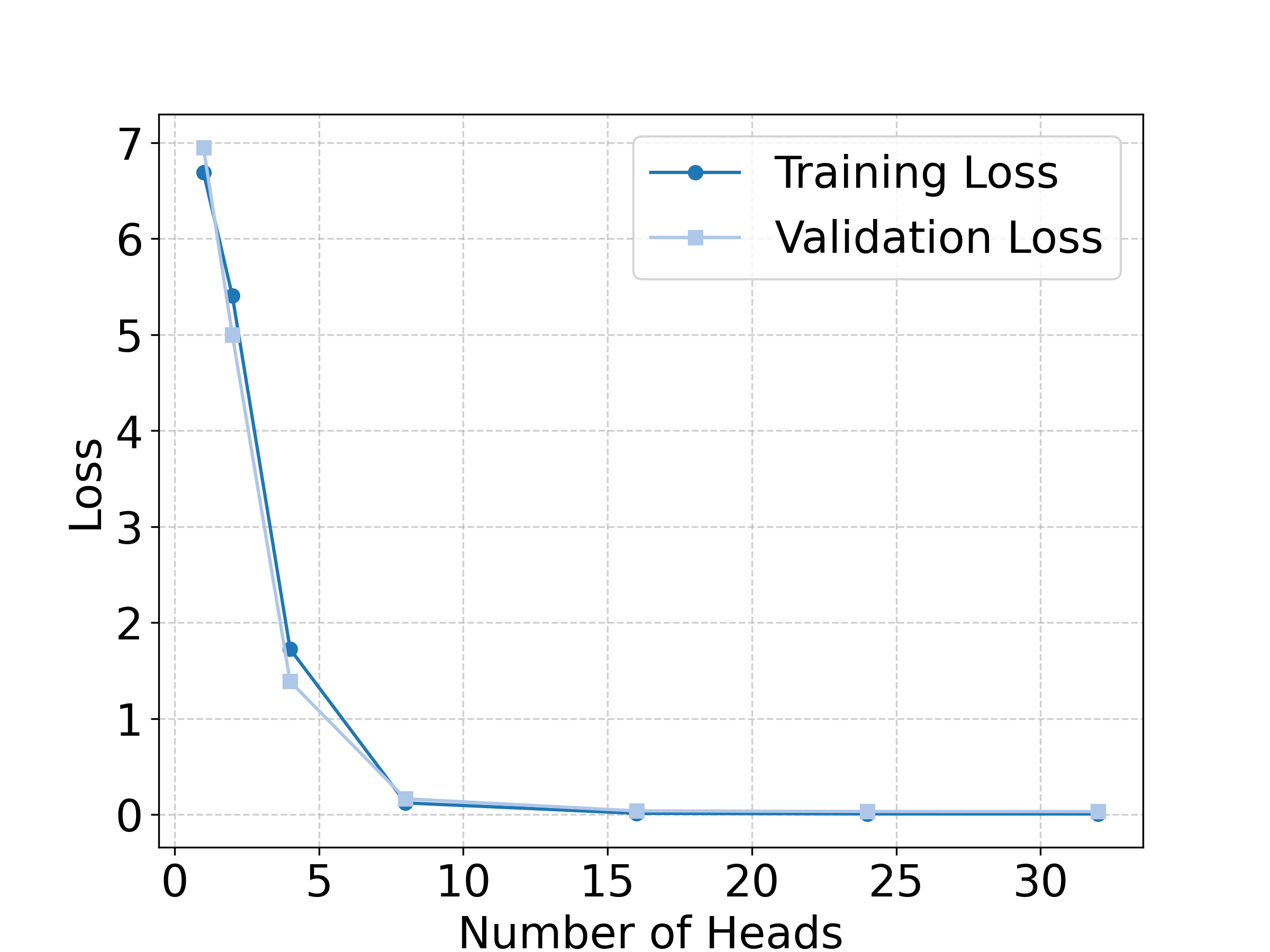} 
    \caption{Loss vs Number of Heads} 
    \label{exp: 6 layer number of heads} 
    \end{minipage}
    \hfill
    \begin{minipage}{0.48\textwidth}
        \centering
        \includegraphics[width=\textwidth]{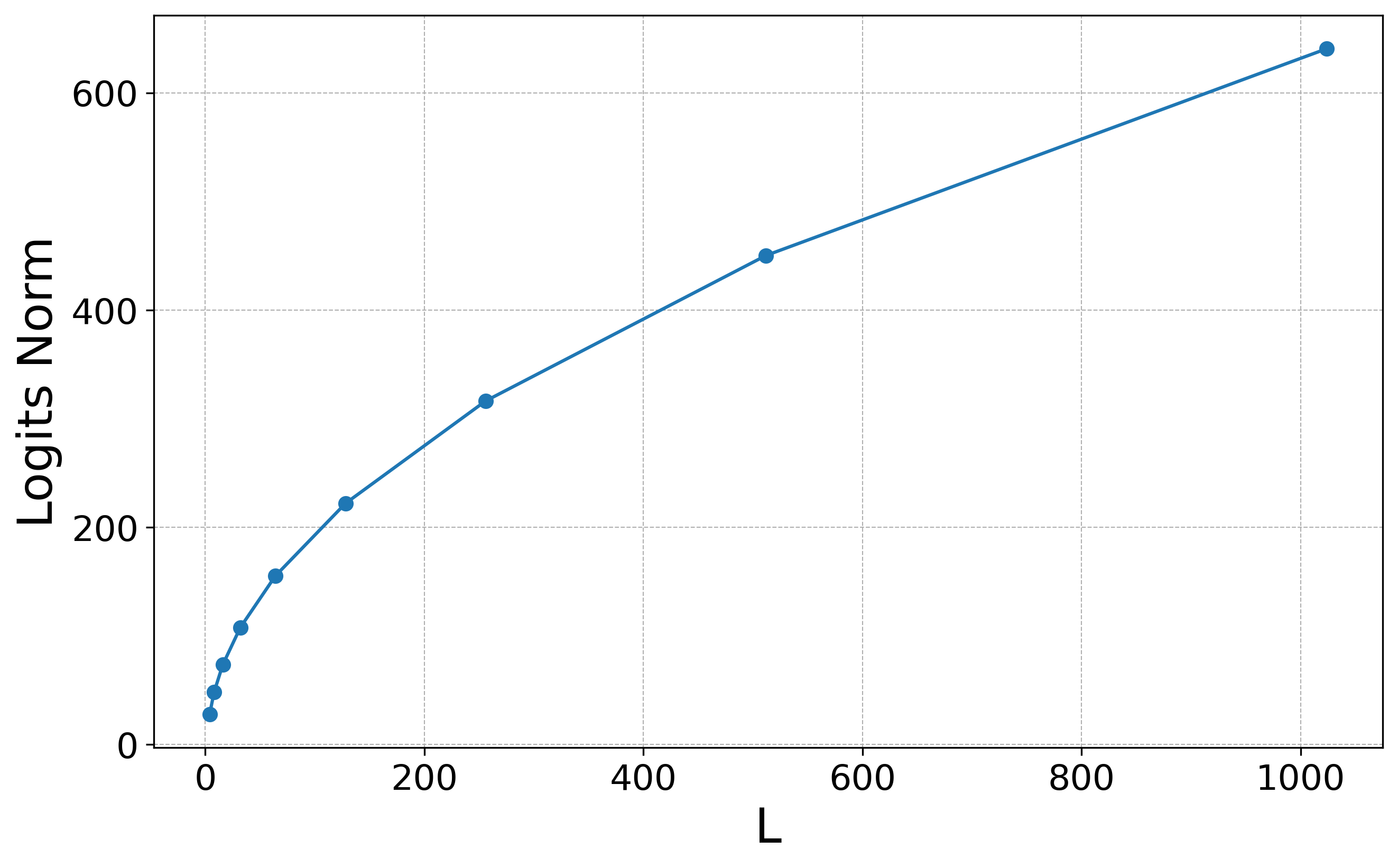} 
    \caption{Logits Norm vs L} 
    \label{exp: logits norm} 
    \end{minipage}
\end{figure}

\subsection{Analysis of Attention Logits Norms}

In Figure \ref{exp: logits norm}, we investigate how the $\ell_2$ norm of attention logits (pre-softmax) varies with sequence length. For each sequence length $L$, we randomly sampled 10 text sequences from the Wikitext-103 dataset and performed forward passes using the \texttt{TinyLlama-1.1B model}. We focused on the logits of the first attention head in the first layer (layer 0, head 0) and computed the $\ell_2$ norm at the middle token of each sequence ($\text{index} = L/2$).

This analysis has several theoretical motivations:  
\begin{itemize}
    \item \textbf{Gradient stability:} The magnitude of logits directly affects the softmax gradients. Excessively large norms may cause gradient explosion, while very small norms lead to overly uniform attention distributions.  
    \item \textbf{Numerical stability:} Extreme logits values can result in numerical overflow, impacting training stability.  
    \item \textbf{Attention concentration:} Differences in logits magnitudes indicate the degree of attention focus, with larger variations corresponding to more concentrated attention.  
\end{itemize}

By analyzing the logits norm at the middle token, we avoid boundary effects and ensure comparability across sequences of different lengths. We hypothesize that the logits norm exhibits a specific scaling behavior with sequence length and should remain relatively stable to support effective gradient propagation. These observations provide insights into the internal mechanisms of Transformers when handling long sequences and can guide design choices for long-sequence modeling.

\subsection{Controlled Ablations for the Two-stage Reduction Pipeline}
\label{appendix: two stage ablation}
We conduct controlled ablations on the corrected permanent low-rank implementation. The reported sweep uses seed 42. The main comparison fixes layers 5--12 and $d_h=16$, while all other settings remain unchanged.

\begin{table}[htbp]
\centering
\caption{Controlled ablation at layers 5--12 and $d_h=16$. WikiText-2 losses are evaluated immediately after initialization and after Stage I; SST-2 accuracy is evaluated after Stage II (or without Stage II when it is disabled).}
\label{tab: two-stage-component-ablation}
\begin{tabular}{lccccc}
\toprule
Initialization & Stage I & Stage II & Init. loss & Stage-I loss & SST-2 Acc. (\%) \\
\midrule
SVD & \checkmark & \checkmark & 5.072 & 3.215 & 93.81 \\
Kaiming & \checkmark & \checkmark & 7.180 & 3.582 & 91.74 \\
SVD & $\times$ & \checkmark & 5.072 & 5.072 & 91.86 \\
SVD & \checkmark & $\times$ & 5.072 & 3.215 & 47.48 \\
\bottomrule
\end{tabular}
\end{table}

\begin{table}[htbp]
\centering
\caption{SST-2 validation accuracy (\%) for Kaiming initialization + Stage I + Stage II.}
\label{tab: ablation-kaiming}
\begin{tabular}{lrrrrrr}
\toprule
Layers $\backslash$ $d_h$ & 2 & 4 & 8 & 16 & 32 & 64 \\
\midrule
1--8 & 77.87 & 79.93 & 83.03 & 86.81 & 87.96 & 88.30 \\
5--12 & 88.65 & 91.06 & 91.63 & 91.74 & 92.66 & 92.78 \\
9--16 & 93.35 & 93.23 & 93.23 & 93.58 & 93.00 & 94.61 \\
Odd layers & 81.77 & 84.63 & 88.19 & 89.91 & 90.37 & 90.83 \\
1--16 & 78.10 & 80.16 & 82.45 & 84.40 & 87.16 & 88.07 \\
\bottomrule
\end{tabular}
\end{table}

\begin{table}[H]
\centering
\caption{SST-2 validation accuracy (\%) for SVD initialization + Stage II, without Stage I.}
\label{tab: ablation-no-stage1}
\begin{tabular}{lrrrrrr}
\toprule
Layers $\backslash$ $d_h$ & 2 & 4 & 8 & 16 & 32 & 64 \\
\midrule
1--8 & 71.79 & 79.93 & 81.65 & 86.70 & 89.91 & 94.50 \\
5--12 & 89.68 & 91.28 & 91.63 & 91.86 & 94.04 & 94.84 \\
9--16 & 91.86 & 93.69 & 94.38 & 94.38 & 94.95 & 95.07 \\
Odd layers & 82.22 & 88.88 & 90.14 & 91.63 & 93.12 & 94.84 \\
1--16 & 76.03 & 81.77 & 84.06 & 87.73 & 90.60 & 95.53 \\
\bottomrule
\end{tabular}
\end{table}

\begin{table}[H]
\centering
\caption{SST-2 validation accuracy (\%) for SVD initialization + Stage I, without Stage II.}
\label{tab: ablation-no-stage2}
\begin{tabular}{lrrrrrr}
\toprule
Layers $\backslash$ $d_h$ & 2 & 4 & 8 & 16 & 32 & 64 \\
\midrule
1--8 & 48.28 & 49.89 & 47.94 & 47.59 & 47.71 & 48.51 \\
5--12 & 48.39 & 47.82 & 48.85 & 47.48 & 47.71 & 48.74 \\
9--16 & 49.54 & 48.28 & 50.11 & 48.51 & 48.51 & 49.08 \\
Odd layers & 46.22 & 47.36 & 46.79 & 47.94 & 48.28 & 49.08 \\
1--16 & 47.25 & 48.74 & 47.13 & 47.94 & 47.13 & 48.28 \\
\bottomrule
\end{tabular}
\end{table}







\end{document}